\documentclass[dvipsnames]{article} %
\usepackage{colm2024_conference}
\pdfoutput=1
\usepackage{booktabs}
\usepackage{graphicx}
\usepackage{enumitem}
\usepackage{wrapfig}
\usepackage{algorithm}
\usepackage{algpseudocode}
\usepackage{microtype}
\usepackage{amsmath}
\usepackage{amsthm}
\usepackage{colortbl}
\usepackage[utf8]{inputenc}
\definecolor{lightgray}{rgb}{0.9,0.9,0.9}
\usepackage{caption}
\usepackage{subcaption}
\usepackage{setspace}
\usepackage{url}
\usepackage{multirow}
\usepackage{colortbl}
\usepackage{tabularx}
\usepackage{blindtext}
\usepackage{pgfplots}
\pgfplotsset{compat=1.18} 
\usepackage{tikz}
\usetikzlibrary{er,positioning,bayesnet}
\usepackage{makecell}
\usepackage{tipa}
\usepackage{siunitx}
\usepackage{nicefrac}
\usepackage{tocloft}
\usepackage{listings}
\usepackage{cleveref}
\usepackage{placeins}
\usepackage{breakcites}
\usepackage[raster,skins]{tcolorbox} %
\usepackage{xltabular}
\usepackage{adjustbox}
\usepackage{xurl}
\usepackage{rotating}
\usepackage[normalem]{ulem}
\useunder{\uline}{\ul}{}
\usepackage{CJKutf8}
\usepackage{pifont}
\usepackage{xcolor}

\usepackage{amsmath,amsfonts,bm}

\def\eqref#1{equation~\ref{#1}}

\def\1{\bm{1}}

\DeclareMathAlphabet{\mathsfit}{\encodingdefault}{\sfdefault}{m}{sl}
\SetMathAlphabet{\mathsfit}{bold}{\encodingdefault}{\sfdefault}{bx}{n}

\newcommand*\justify{%
  \fontdimen2\font=0.4em%
  \fontdimen3\font=0.2em%
  \fontdimen4\font=0.1em%
  \fontdimen7\font=0.1em%
  \hyphenchar\font=`\-%
}

\renewcommand{\texttt}[1]{%
  \begingroup
  \ttfamily
  \begingroup\lccode`~=`/\lowercase{\endgroup\def~}{/\discretionary{}{}{}}%
  \begingroup\lccode`~=`[\lowercase{\endgroup\def~}{[\discretionary{}{}{}}%
  \begingroup\lccode`~=`.\lowercase{\endgroup\def~}{.\discretionary{}{}{}}%
  \catcode`/=\active\catcode`[=\active\catcode`.=\active
  \justify\scantokens{#1\noexpand}%
  \endgroup
}

\newcommand{\ourmodel}{\emph{ERNIE 5.0}\xspace}
\newcommand{\basemodel}{\emph{ERNIE 5.0-Base}\xspace}

\title{\ourmodel{} Technical Report}

\author{
\bf ERNIE Team, Baidu \\
\vspace{0.1cm}
\texttt{ernie@baidu.com}
}

\makeatletter\def\@abstract{
In this report, we introduce \ourmodel, a natively autoregressive foundation model desinged for unified multimodal understanding and generation across text, image, video, and audio.
All modalities are trained from scratch under a unified next-group-of-tokens prediction objective, based on an ultra-sparse mixture-of-experts (MoE) architecture with modality-agnostic expert routing.
To address practical challenges in large-scale deployment under diverse resource constraints, \ourmodel adopts a novel elastic training paradigm.
Within a single pre-training run, the model learns a family of sub-models with varying depths, expert capacities, and routing sparsity, enabling flexible trade-offs among performance, model size, and inference latency in memory- or time-constrained scenarios.
Moreover, we systematically address the challenges of scaling reinforcement learning to unified foundation models, thereby guaranteeing efficient and stable post-training under ultra-sparse MoE architectures and diverse multimodal settings.
Extensive experiments demonstrate that \ourmodel achieves strong and balanced performance across multiple modalities.
To the best of our knowledge, among publicly disclosed models, \ourmodel represents the first production-scale realization of a trillion-parameter unified autoregressive model that supports both multimodal understanding and generation.
To facilitate further research, we present detailed visualizations of modality-agnostic expert routing in the unified model, alongside comprehensive empirical analysis of elastic training, aiming to offer profound insights to the community.
}\makeatother

\begin{document}

\maketitle

\newpage

\clearpage
\pagestyle{firstpage}  %
\tableofcontents
\clearpage
\pagestyle{normalpage}

\section{Introduction}
\label{sec:intro}

Recent advances in large language and vision-language models, including ERNIE~\citep{ernie45tech}, Gemini~\citep{gemini2.5,gemini3.0}, GPT~\citep{gpt4o,gpt5}, Claude~\citep{claude}, DeepSeek~\citep{deepseekv3,deepseekv3.2}, and Qwen~\citep{qwen3,qwen3-vl}, demonstrate that large-scale autoregressive sequence modeling provides a powerful foundation for language and multimodal understanding.
By modeling diverse inputs as sequences of tokens, these models exhibit strong reasoning and alignment capabilities across modalities.
However, in most existing systems, autoregressive modeling serves multimodal understanding while the output is still text-centric, which restricts the model’s ability to engage in various multimodal interactions.
To overcome this limitation, recent approaches augment pre-trained language models with modality-specific decoders or generators, which are connected to the language backbone through late-fusion designs~\citep{qwen3-omni,seedream4}.
Although effective for individual modalities, these designs decouple multimodal generation from understanding and rely on modality-specific, non-autoregressive objectives, which hinder deep cross-modal integration and often force a trade-off between multimodal integration and core language performance.
Against this backdrop, designing a unified autoregressive paradigm remains a prominent open challenge. Such a framework  must natively support both multimodal understanding and generation, preserve strong unimodal capabilities, and scale effectively as  model and data sizes continue to grow.

In this report, we introduce \ourmodel, a next-generation foundation model natively designed to integrate text, image, audio, and video capabilities \emph{under a unified autoregressive framework} for both multimodal understanding and generation.
Rather than augmenting a pre-trained language model with modality-specific components, \ourmodel trains all modalities simultaneously \emph{from scratch}, which alleviates the ``ability seesaw'' problem observed in later-fusion approaches and ensures that all modalities evolve collectively without sacrificing performance.
Specifically, heterogeneous inputs are mapped into a shared token space, and modeling across all modalities is formulated under a unified \emph{Next-Group-of-Tokens Prediction} objective, which avoids explicit modality boundaries and inconsistent optimization trajectories.
To support scalability, \ourmodel leverages an ultra-sparse Mixture-of-Experts (MoE) backbone with \emph{modality-agnostic expert routing}. 
Routing decisions are conditioned on unified token representations rather than modality identifiers, allowing tokens from various modalities to be dispatched to a shared pool of experts.
This ultra-sparse, modality-agnostic architecture eliminates the need for heuristic modality-specific expert allocation, offering sufficient capacity for both differentiation and collaboration among modality-specialized behaviors.

During pre-training of \ourmodel, we propose a novel elastic training paradigm that enables a single pre-training run to produce a family of models with varying capacity–efficiency trade-offs.
Instead of optimizing a static architecture, our elastic training approach dynamically samples sub-models with varying depth, width, and routing sparsity for each training instance, guided by a pre-defined schedule. Both the sampled sub-models and the full-model are optimized in one backpropagation process under the same autoregressive objective.
The sub-model sampling strategy improves the functional integrity of parameters and maintains competitive performance even when only a subset of parameters is available.
Elasticity in depth and width facilitates the production of models with smaller sizes, whereas the sparsity elasticity reduces the number of activated experts during inference, leading to higher throughput and improved computational efficiency.
Meanwhile, elastic training enables sub-models to inherit knowledge from the full model and provides flexible instantiation of smaller models in subsequent post-training stages, thereby eliminating the need to pre-train multiple models of various sizes or rely on customized compression, and making \ourmodel well suited for deployment under diverse hardware, memory, and latency constraints. 

Following unified pre-training, we conduct a multi-stage post-training pipeline that combines supervised fine-tuning (SFT) with unified multimodal reinforcement learning (UMRL).
The coexistence of heterogeneous multimodal inputs and ultra-sparse MoE architecture introduces substantial optimization challenges, which increases the sensitivity of UMRL to sampling bias, sparse reward signals, and entropy collapse.
To cope with these issues, we build a unified verifier system and develop a suite of scalable techniques to improve the stability and efficiency of RL training.
An unbiased replay buffer is employed to improve rollout efficiency while preserving a balanced data distribution. 
Multi-granularity importance sampling, together with positive sample masking, stabilizes policy optimization and effectively mitigates entropy collapse.
For difficult tasks with sparse rewards, adaptive hint-based RL is introduced to provide auxiliary guidance when needed.
By ensuring stable and efficient post-training, these designs support the excellent multimodal reasoning ability in \ourmodel.

For infrastructure, we utilize hybrid parallelism with fine-grained memory control to support effective training of a trillion-parameter ultra-sparse MoE model.
For unified multimodal training, tokenizers are decoupled from the MoE backbone and deployed on separate GPU nodes, so that each component can adopt its most suitable parallelization strategy.
To accommodate local bidirectional attention in vision, we employ FlashMask~\citep{flashmask} to efficiently handle per-sample heterogeneous attention masks.
Finally, we design a scalable and disaggregated RL infrastructure that coordinates training, rollout, and environment interaction, to ensure high-throughput and numerically consistent post-training.

We evaluate \ourmodel on a diverse set of text and multimodal benchmarks spanning perception, reasoning, understanding, and generation.
Across these tasks, \ourmodel consistently matches or outperforms specialized baselines, indicating that unified training retains strong modality-specific performance without architectural fragmentation.
Ablation results further highlight the effectiveness of modality-agnostic expert routing and elastic training.
Despite employing a single shared routing mechanism across modalities, experts exhibit clear specialization patterns that are primarily shaped by task requirements rather than modality boundaries.
Reducing routing top-$k$ to 25\% during inference yields over 15\% decoding speedup with only minor accuracy loss, while elastic training across depth, width, and sparsity preserves near-full performance using only 53.7\% activated parameters and 35.8\% total parameters, suggesting a scalable and efficient foundation for next-generation unified multimodal models.

In the following sections, we systematically present the design of the model architecture and its core technical components, followed by a detailed description of the training and optimization pipeline. 
We then evaluate \ourmodel on a comprehensive set of benchmarks to validate the effectiveness of the proposed unified framework. Finally, we share some key technical insights during the model training process, hoping to be helpful for future research on scalable and general-purpose foundation models.

\section{Architecture}

\begin{figure}[t]
  \centering
    \includegraphics[width=0.95\linewidth]{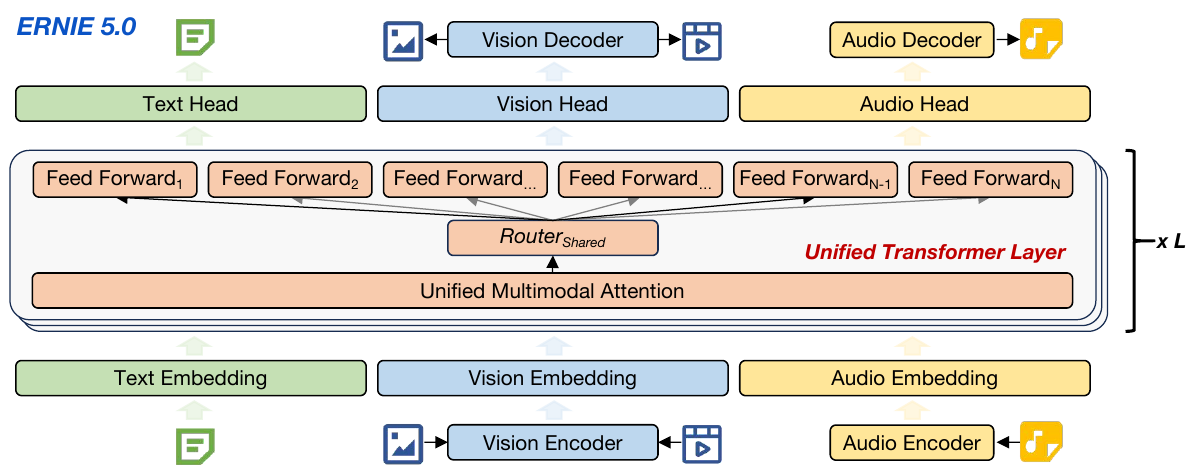} 
    \caption{
\ourmodel architecture. It is trained from scratch under a unified autoregressive paradigm that integrates multimodal understanding and generation.
Text, vision, and audio are encoded and serialized, then processed by a unified backbone.
An ultra-sparse MoE architecture with modality-agnostic routing is employed, in which tokens from different modalities are dispatched to a shared expert pool.
}
\label{fig:arch}
\end{figure}

As shown in Figure~\ref{fig:arch}, \ourmodel adopts an ultra-sparse mixture-of-experts architecture that integrates language, image, video, and audio within a single autoregressive framework for both multimodal understanding and generation.
The model consists of a shared backbone for unified sequence modeling, together with visual and audio tokenizers that convert multimodal inputs into a unified token sequence.
All modalities are trained under a shared \emph{Next-Group-of-Tokens Prediction} objective, enabling deep cross-modal interactions with end-to-end optimization.
In this section, we first describe the autoregressive backbone in Sec.~\ref{sec:backbone}, which serves as the core of \ourmodel, followed by the visual and audio processing pipelines in Secs.~\ref{sec:vision} and~\ref{sec:audio}.

\subsection{Unified Autoregressive Backbone with Ultra-Sparse Mixture-of-Experts}
\label{sec:backbone}

Heterogeneous modalities differ substantially in token semantics, temporal structures, and optimization dynamics, making naive cross-modal parameter sharing prone to unstable optimization and performance degradation, especially when modeling understanding and generation in a single model.

To address these challenges, \ourmodel is designed to be trained from scratch under a unified autoregressive framework.
Multimodal inputs, including text, image, video, and audio, are projected into a shared token space, serialized into a unified sequence, and optimized under the \emph{Next-Group-of-Tokens Prediction} objective. 
Specifically,  
text generation adheres to the standard \emph{Next-Token Prediction} (NTP) paradigm, augmented by the Multi-Token Prediction (MTP) mechanism~\citep{gloeckle2024better, deepseekv3}  to enhance both output quality and inference efficiency.
For vision and audio modalities, generation is formulated as a group-of-tokens prediction task, so as to align their generative processes with the text autoregressive objective.
Vision generation employs \emph{Next-Frame-and-Scale Prediction} (NFSP)~\citep{ji2026videoar}, and audio generation utilizes \emph{Next-Codec Prediction} (NCP) to capture temporal and spectral structure.
By unifying heterogeneous modalities under a single optimization objective, all tokens are trained within a consistent sequence prediction paradigm, enabling training from scratch and deep token-level multimodal interactions throughout the unified backbone.

Although unifying the learning paradigm helps reduce modality discrepancies, it cannot fully eliminate the intrinsic differences across modalities, necessitating substantial model capacity to capture diverse multimodal knowledge.
To this end, \ourmodel adopts a sparse Mixture-of-Experts (MoE) architecture to scale model capacity efficiently while controlling both training and inference costs.
At the core of this architecture is \emph{modality-agnostic expert routing}, where routing decisions are conditioned on unified token representations rather than explicit modality identifiers.
The router dispatches tokens from different modalities to a shared pool of experts, enabling effective cross-modal parameter sharing.
In contrast to modality-isolated routing strategies in our previous models~\citep{ernie45tech}, such a unified routing mechanism promotes cross-modal knowledge generalization and improves single-modality performance through the emergent specialization of shared experts.
Moreover, it obviates the requirement of heuristic modality-specific expert allocation, which is often non-trivial in practice, especially when more than two modalities are involved.
By employing an ultra-sparse and fine-grained MoE architecture,  \ourmodel achieves an activation rate below 3\%, allowing the model to substantially expand its effective capacity without incurring a proportional increase in computational overhead.
The training is further stabalized by an auxiliary-loss-free load balancing~\citep{wang2024auxiliary} , ensuring  robust expert utilization at a trillion-parameter scale.

Based on the unified optimization objective and the shared parameter space, \ourmodel formally integrates multimodal understanding and generation within a single autoregressive backbone.
Despite such formal unification, some challenges remain in learning representations that can convincingly support both tasks.
Typically, multimodal understanding focuses on abstract and semantic-level concepts, whereas generation requires accurate modeling of fine-grained perceptual details.
\ourmodel is therefore designed to learn a unified representation that captures high-level semantics while preserving fine-grained details, enabling both comprehension and synthesis in a unified manner.
In the unified framework, semantic-level signals guide generative modeling toward global consistency, while generative training, in turn, strengthens fine-grained perception and detail-sensitive reasoning.
This mutual reinforcement allows a single backbone to robustly support perception, reasoning, and creative generation.
Based on this design philosophy, we further introduce unified visual and audio input–output interfaces and their corresponding processing pipelines in the following sections, which constitute a distinctive feature of \ourmodel compared to previous models.

\subsection{Visual Modeling}
\label{sec:vision}

\begin{figure}[t]
  \centering
    \includegraphics[width=0.9\linewidth]{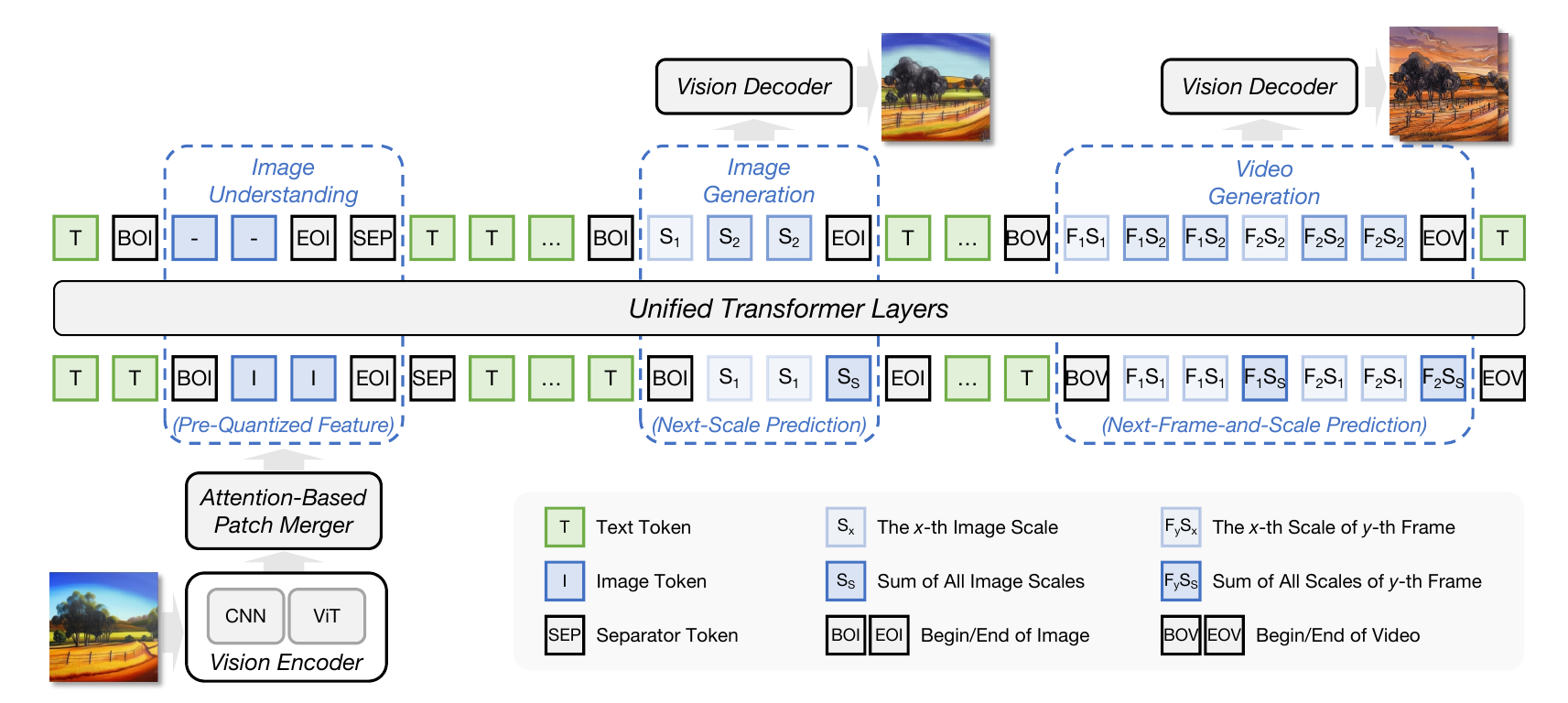} 
    \caption{
Overview of the unified vision understanding and generation architecture.
For understanding, visual features are extracted by a hybrid CNN–ViT representation and then compressed via an Attention-Based Patch Merger.
For generation, we introduce the Next-Frame-and-Scale Prediction (NFSP) paradigm, where image generation is formulated as Next-Scale Prediction, and video generation further extends this process with Next-Frame prediction along the temporal dimension.
}
\label{fig:arch_vision}
\end{figure}

In \ourmodel, image is treated as a special case of video (e.g., a single-frame video), sharing the unified design philosophy of visual understanding and generation.
Visual understanding is built upon a hybird representation that encodes both global semantic information and local perceptual details, allowing high-level reasoning while preserving fine-grained visual sensitivity.
For image and video generation, a visual autoregressive paradigm is proposed to support coherent modeling across both spatial and temporal dimensions in discrete token space, as shown in Figure~\ref{fig:arch_vision}.
In the following, we first introduce vision tokenization, followed by our tailored designs for visual understanding and generation.

\subsubsection{Vision Tokenization}
\label{sec:vision_tokenization}

To support autoregressive visual modeling across both spatial and temporal dimensions, \ourmodel propose \emph{Next-Frame-and-Scale Prediction} (NFSP), where image generation is formulated as a \emph{Next-scale Prediction} problem, and video generation further extends this formulation with \emph{Next-Frame Prediction}~\citep{ji2026videoar}.
To this end, we first train a causal 2D multi-scale tokenizer for images, which provides strong spatial representations through large-scale image pre-training. Building upon this image tokenizer, we inflate it into a causal 3D convolutional tokenizer, thereby unifying image and video tokenization within a single model.
The progressive design preserves the spatial modeling capabilities learned from images while introducing temporal perception for videos, leading to faithful reconstruction of high-level visual elements such as scene text and human faces.

During tokenizer training, we incorporate auxiliary supervision signals to enhance representation quality and training stability. 
Specifically, we utilize the adversarial loss~\citep{stylegan} from GAN-based discriminators to improve distributional fidelity.
Meanwhile, we incorporate a semantic branch and apply a semantic regularization loss derived from large-scale vision foundation models to preserve high-level semantic consistency.
These complementary objectives improve the learnability and stability of visual tokens, as well as facilitate effective autoregressive modeling in the unified backbone.

Following the bit-wise quantization strategy, we quantize the unified visual latent representation into a group of bit-codes, where the number of bits directly corresponding to the size of discrete vocabulary~\citep{han2025infinity}.
Based on this mechanism, we pre-train a series of tokenizers with progressively increasing bit numbers.
During the training of \ourmodel, we adopt a progressive tokenizer switching strategy, starting with a low-bit tokenizer (i.e., small vocabulary) and gradually transitioning to higher-bit variants (i.e., larger vocabularies).
By first learning coarse-grained, low-bit representations with small vocabularies and progressively introducing finer-grained, higher-bit tokenizers, the backbone follows a smoother and stable optimization trajectory, effectively alleviating early-stage training instability and leading to improved visual generation quality.

\subsubsection{Visual Understanding with Dual-Path Hybrid Representation}

While the above design mainly targets visual tokenization for unified autoregressive modeling and generation, we observe that, prior to quantization, visual features are typically compressed by a downsampling module into low-dimensional representations, whose dimensionality is aligned with the tokenizer bit-width.
The compression inevitably leads to the loss of fine-grained semantic information, which has also been widely observed to limit the performance of visual understanding tasks~\citep{unitok2025}.

To address this issue, \ourmodel directly leverages the dual-path visual features prior to quantization. 
We integrate perceptual features extracted by Convolutional Neural Networks (CNNs) with semantic features encoded by a Vision Transformer (ViT).
However, the representations produced by these two paths are misaligned in spatial structure, our empirical study shows that roughly fusing CNN and ViT features through MLP-based adapters often fails to fully exploit their complementary strengths and introduces representational interference, resulting in degraded understanding performance. 
This observation motivates the following Attention-based Patch Merger.

Formally, given a spatial token in an image or a spatio-temporal token in a video, we extract two sets of features, 
$\mathbf{F}_{cnn} \in \mathbb{R}^{N \times K \times D_{cnn}}$ and
$\mathbf{F}_{vit} \in \mathbb{R}^{N \times K \times D_{vit}}$, where $N$ is the number of visual understanding tokens and $K$ is the number of local patches grouped for each token, $D_{cnn}$ and $D_{vit}$ are the feature dimensions.
In image understanding tasks, we group $K=4$ spatially adjacent patches, while in video understanding tasks, we group $K=16$ patches spanning $4$ neighboring frames.
Before feature fusion, we project CNN features to match the ViT feature space, and then concatenate the aligned CNN and ViT patch features along the patch dimension to obtain $\mathbf{F}_{mrg} \in \mathbb{R}^{N \times 2K \times D_{vit}}$. 
Next, multi-head self-attention is applied to the concatenated patch tokens, $\mathbf{Z} = \mathrm{Attn}(\mathbf{F}_{mrg})$, where the attention mechanism jointly models correlations between CNN and ViT features, as well as spatial and temporal dependencies among the group of patches. 
The output preserves the same shape, $\mathbf{Z} \in \mathbb{R}^{N \times 2K \times D_{vit}}$.
Finally, we perform mean pooling over the patch dimension to obtain a compact representation $\mathbf{F}_{out} \in \mathbb{R}^{N \times D_{vit}}$, which is then projected to align with the embedding dimension of the unified backbone.

The choice of feature fusion has a significant impact on model performance, and the naive MLP-based fusion is proven inadequate to effectively integrate CNN and ViT features.
In contrast, the proposed attention-based aggregation module consistently outperforms both CNN-only and ViT-only baselines on a wide range of benchmarks without introducing noticeable computational overhead, with particularly pronounced gains in document and chart understanding as well as general visual understanding tasks.
Importantly, \ourmodel is designed as a unified framework for visual understanding and generation, where visual representations must support not only discriminative tasks but also fine-grained generative tasks such as pixel-level image and video editing. 
By utilizing attention, the module adaptively aggregates local patches together with high-level semantic information, capturing critical visual features while reducing the number of visual tokens. It results in a compact yet expressive visual representation that provides a strong and stable foundation for various vision-language tasks.

\subsubsection{Visual Generation with Next-Frame-and-Scale Prediction}

Recall that the Next-Frame-and-Scale Prediction (NFSP) paradigm introduced in vision tokenization formulates visual generation in an autoregressive manner, where image generation is viewed as a special case of single-frame video generation.
Under this formulation, the model predicts visual tokens across multiple spatial scales within each image (or each frame) for image generation, while performing frame-wise prediction along the temporal dimension for video generation.
When predicting tokens at a certain scale, the model takes the previous generated scales as input, and a \emph{scale-wise causal attention mask} is applied, where tokens within the current scale are bidirectionally visible and are predicted in parallel, while tokens from all previous scales and historical frames are visible in a causal (uni-directional) manner. 
The NFSP paradigm disentangles spatial and temporal modeling, that is, intra-frame prediction from low-resolution to high-resolution captures fine-grained spatial structures, whereas next-frame prediction models inter-frame temporal dependencies.

To support positional modeling of heterogeneous tokens across spatial and temporal dimensions, we introduce a \emph{Unified Spatiotemporal Rotary Positional Embedding (Uni-RoPE)} and apply it to all tokens in \ourmodel.
For a unified sequence of length $N$, the positional encoding of the $i$-th token is defined as $\mathrm{Uni}$-$\mathrm{RoPE}_i = (t_i, h_i, w_i), i \in \{1, \ldots, N\}$.
For text and audio tokens, we set $t_i = h_i = w_i$, where the shared value follows token index in the sequence.
For visual tokens, $t_i$ is used for frame indexing, which increases monotonically to preserve temporal ordering, and $(h_i, w_i)$ corresponds to spatial locations within each frame. 
To ensure spatial consistency across multi-scales, we adopt a center-aligned coordinate strategy, where tokens at different scales are aligned based on the geometric centers of the scale.

Empirically, auto-regressive visual generation is susceptible to error accumulation over extremely long token sequences. 
To mitigate this issue, we corrupt historical tokens during training by \emph{randomly flipping} their bits, while supervising the model to self-correct toward the ground-truth tokens of the current scale. The corruption-based training strategy improves robustness against compounding errors in long-horizon generation.
Meanwhile, we apply a \emph{loss reweighting} strategy to emphasize early-stage predictions and alleviate the token imbalance introduced by multi-scale tokenization.
For video generation in particular, we further introduce \emph{windowed temporal attention} and \emph{random historical frame masking} to encourage the model to focus on relevant temporal context and improve robustness~\citep{ji2026videoar}.

Within the token-based modeling paradigm, training visual generation abilities under a fixed modality token budget poses a fundamental challenge for high-resolution images and videos. 
Increasing visual resolution enlarges the token sequence length, which in turn reduces the effective training batch size and degrades optimization stability.
To address this challenge, we adopt a cascaded diffusion refiner on top of the autoregressive backbone. The backbone generates low-resolution samples with precise semantics and structural layout, while the refiner focuses on enhancing fine-grained visual details at higher resolution.
The diffusion refiner is trained separately from the backbone, using paired low-resolution samples with controlled degradation, together with their corresponding high-resolution images or videos.
The decoupled training scheme enables high-fidelity refinement while preserving the complete semantic and structure produced by the autoregressive model, and avoids optimization conflicts caused by introducing autoregressive and diffusion losses within a shared backbone.

\begin{figure}[t]
  \centering
    \includegraphics[width=0.9\linewidth]{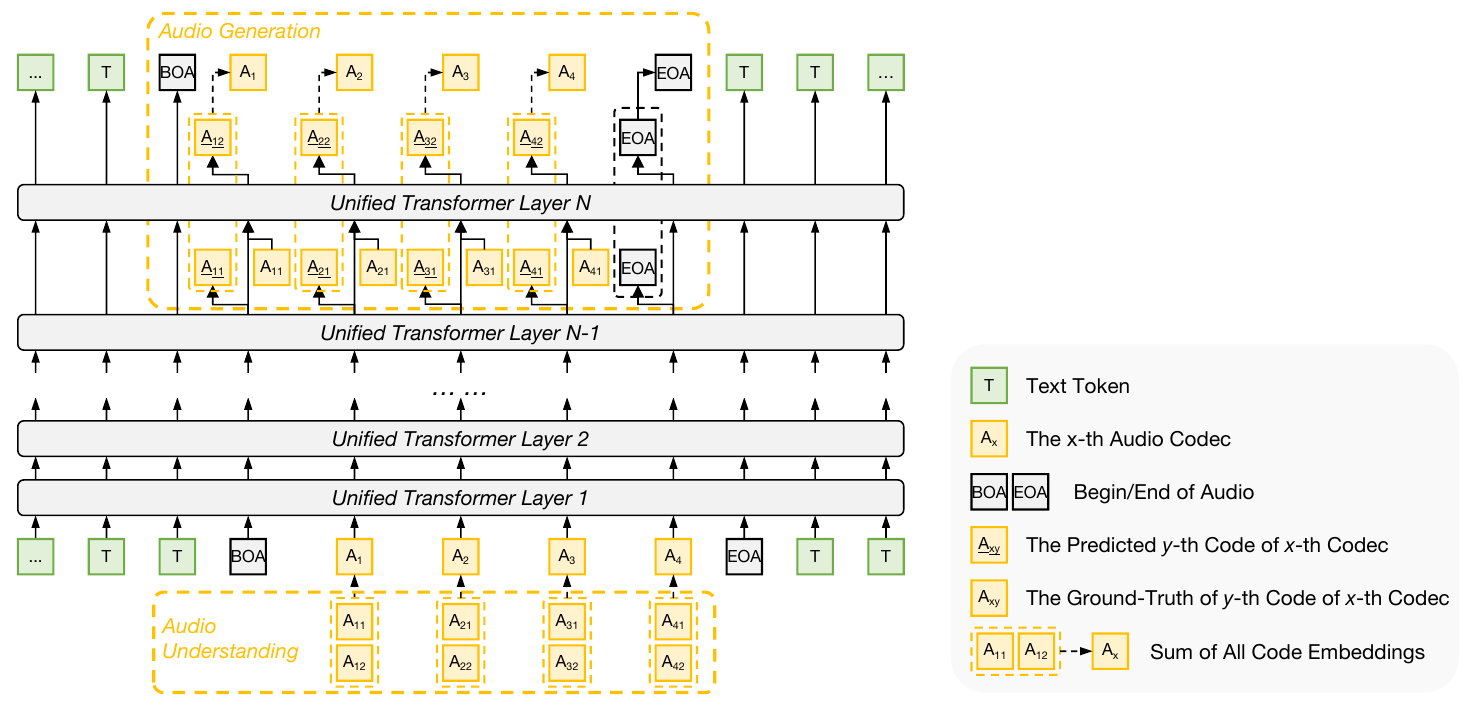} 
    \caption{
Overview of the depth-wise architecture for audio understanding and generation.
For understanding, embeddings from multiple residual levels are additively combined to form audio token representation.
For generation, \ourmodel introduces Next-Codec Prediction (NCP) to achieve hierarchical prediction across transformer layers, where the ground-truth code embedding (or the predicted one during inference) is fed back to condition subsequent predictions.
}
\label{fig:arch_audio}
\end{figure}

\subsection{Audio Modeling}
\label{sec:audio}

Similar to the vision modality, audio modeling in \ourmodel is also formulated under a unified autoregressive, token-based framework that supports understanding and high-fidelity generation.
Inspired by the success of neural audio codecs~\citep{kumar2023high, zhang2024speechtokenizer}, audio signals are represented as hierarchical discrete codec tokens that capture higher-level semantics and fine-grained acoustic details.
To avoid the prohibitive sequence length caused by flattening multi-codebook tokens into a single sequence, we introduce a depth-wise autoregressive architecture that performs structured prediction across codec dimensions, as illustrated in Figure~\ref{fig:arch_audio}.

\subsubsection{Audio Tokenization}

Given an input waveform, we first map the continuous audio signal into a sequence of discrete tokens using a codec-style tokenizer with a token rate of 12.5~Hz.
The audio quantization module follows a Residual Vector Quantization (RVQ) design, decomposing the signal into multiple tokens at different levels of granularity. 
In that case, the first token is explicitly assigned to encode high-level audio semantics, while the remaining tokens encode residual acoustic information with progressively finer details.
To ensure that the first token captures rich semantic information, including linguistic and phonetic cues, required for audio–text modeling, we distill knowledge from a pretrained Whisper model~\citep{radford2023robust}.
Specifically, we align the representation of the first audio token with the encoder outputs of Whisper.
Average pooling is applied to the Whisper representations to match the token rate of our audio tokenizer, which resolves the temporal mismatch between the teacher and student models.
Complementary to the first semantic token, residual acoustic tokens preserve fine-grained characteristics of the audio signal, such as timbre and prosody.
Together, the hierarchical tokenization process disentangles semantic content from acoustic realization and provides a compact audio representation that is integrated naturally into \ourmodel’s unified autoregressive backbone.

\subsubsection{Audio Understanding and Generation with Next-Codec Prediction}

Based on the audio tokenization described above, we utilize a depth-wise autoregression architecture to model audio tokens for both understanding and generation, drawing inspiration from coarse-to-fine prediction paradigms developed in visual generation~\citep{chen2024spark}.
Instead of flattening all residual audio tokens into a single long sequence, \ourmodel distributes the prediction of residual codes across transformer layers.
Each layer models audio information at a specific level of granularity, which allows multi-level audio representations to be efficiently handled within a unified framework.

For audio understanding tasks, text tokens are embedded using standard text embedding layers, while audio tokens are represented through a depth-wise additive embedding mechanism.
Each audio token consists of multiple discrete codes that correspond to different residual levels.
At each level, the code is mapped to an embedding through a level-specific embedding matrices, and embeddings from all levels are summed to form the final codec representation.
The additive aggregation reflects the residual nature of audio representation, where each depth contributes complementary information at different granularities from coarse to fine.
The resulting audio token representations are placed at the corresponding positions in the input sequence and processed uniformly with text tokens by the autoregressive backbone, enabling seamless multimodal understanding.

For audio generation tasks, \ourmodel introduces Next-Codec Prediction (NCP) to generate hierarchical audio tokens in a coarse-to-fine manner.
Multiple audio heads are inserted into the top transformer layers to support depth-wise prediction.
Conditioned on the multimodal context, the model first predicts the first semantic code and then sequentially generates codes for subsequent residual levels.
After each prediction, the generated code is mapped to its corresponding embedding and added back to the hidden state, which then conditions the prediction at the next level.
During training, teacher forcing is applied, and the feedback embedding is derived from the ground-truth code.
Such iterative process continues until all levels are predicted, allowing high-level semantic information to guide the synthesis of increasingly fine-grained acoustic details.
Once the complete set of hierarchical audio codes is obtained, the audio decoder converts them into waveforms.
For speech synthesis, a speaker embedding is inserted as part of the conditioning context to enable controllable voice timbre, guiding acoustic realization without altering deep semantic content or depth-wise prediction structure.
Overall, the NCP formulation is compatible with the residual design used during audio understanding, which ensures structural alignment between audio input and output.

\section{Pre-Training}
\label{sec:pre}

The pre-training phase forms the foundation of \ourmodel, where the model learns generalizable representations across multiple modalities. 
In this section, we first describe the composition of pre-training data (Sec.~\ref{sec:pretrain-data}), followed by a part of training recipe (Sec.~\ref{sec:pretrain-recipe}). 
Finally, we focus on the \emph{Elastic Training} technique (Sec.~\ref{sec:elastic-training}) introduced in \ourmodel, which enables the production of multiple models of different sizes within a single pre-training run, significantly reducing the computational costs associated with training a series of models while maintaining both efficiency and performance.

\subsection{Pre-Training Data}
\label{sec:pretrain-data}

\ourmodel is trained on an large, high‑quality multimodal dataset that reflects its natively omni design and dual capabilities for both understanding and generation. 
Unlike conventional late-fusion approaches, 
\ourmodel is simultaneously exposed to text, images, videos, and audios from the very beginning of training.
The unified training paradigm enables the model to learn representations that integrate semantic information across all modalities,  while also requiring a large amount of pre-training data to support learning from scratch for each modality.
To manage such diverse data, we build a standardized platform and organize all data according to their input and output modalities.
Based on this organization, the pre-training data are broadly categorized into two groups: text data and multimodal data.

\paragraph{Text Data} The textual component spans a vast collection of multilingual web crawls, curated corpora, books, scientific publications, code repositories, and structured knowledge sources selected for breadth, diversity, and linguistic richness. 
We retrain the text tokenizer to better support large-scale multilingual modeling. Specifically, we encode text in UTF-16BE to provide stable byte-level fallback and a more compact representation for many non-Latin symbols, improving data throughput in multilingual training, and we use BPE dropout~\citep{provilkov2020bpe}  to reduce overfitting to frequent patterns.
It is worth noting that, for languages without explicit whitespace word boundaries (e.g., Chinese), we filter out long unspaced phrases that can be decomposed by standard word-segmentation tools, which helps reduce vocabulary sparsity, improve training efficiency, and enhance model generalization.

\paragraph{Multimodal Data} For visual and audio modalities, we curate a dataset comprising paired image–text, video–text, audio-text, as well as diverse interleaved multimodal sequences where text is integrated with images, videos, and audio, all accompanied by metadata and captions. 
Such data composition connects textual concepts with visual and audio contexts across both spatial and temporal dimensions.
By explicitly modeling cross-modal alignment, the model is able to learn semantic relationships not only within individual modalities and across disparate ones, which supports a wide range of tasks from multimodal understanding to creative multimodal generation.

Rigorous preprocessing and quality controls are applied at scale to maintain both signal integrity and diversity. 
Heuristic and model-based filters remove low-quality and unsafe content, extensive deduplication prevents memorization artifacts, and decontamination safeguards keep benchmarks out of the training data. 
The finalized pre‑training corpus comprises trillions of text tokens and multimodal instances that balance scale with high‑fidelity semantic content. The large, diverse and well‑filtered dataset is fundamental to \ourmodel's strong performance on text and multimodal understanding, reasoning, and generation benchmarks.

\subsection{Training Recipe}
\label{sec:pretrain-recipe}

\ourmodel is trained with a carefully designed recipe to ensure training stability, scalability, and efficient utilization of compute resources. 
The training follows a multi-stage pre-training strategy to  progressively extend context length while maintaining stable optimization dynamics. 

\paragraph{Stage 1: 8K Pre-Training}
The initial stage uses a maximum context length of 8K tokens. We adopt a Warmup-Stable-Decay (WSD) learning rate schedule~\citep{hu2024minicpm} in this stage. The learning rate is linearly warmed up for 2{,}000 steps from zero to a peak value of $1 \times 10^{-4}$, and then kept constant for the remainder of the 8K training stage.
To improve training efficiency and stability at scale, we employ a batch size scheduling strategy, where the global batch size is gradually increased from 14M tokens to 56M tokens during early training.
To facilitate seamless long-context extension, the RoPE base is set to 1{,}000{,}000 starting from the 8K stage. This design choice avoids the need for reparameterization or interpolation during subsequent context length expansion, ensuring lossless and stable long-context training.

\paragraph{Stage 2: 32K\&128K Mid-Training}
In the mid-training stage, we progressively extend the context length to 32K and 128K tokens while keeping the global batch size unchanged. During this phase, we switch to a cosine learning rate schedule and anneal the learning rate from $1 \times 10^{-4}$ to $1 \times 10^{-5}$. 

For MoE-specific optimization, the bias update speed for
auxiliary-loss-free load balancing~\citep{wang2024auxiliary} is set to $1 \times 10^{-4}$ in the 8K pre-training stage and reduced to $1 \times 10^{-5}$ during mid-training, which effectively suppresses iteration-level oscillations observed in large-scale MoE training. 
The MTP loss weight~\citep{deepseekv3} is decreased from 0.3 in the 8K stage to 0.1 during mid-training, ensuring stable adaptation as the model scales to longer contexts.
Besides, we introduce a posterior-based loss weighting strategy that rescales the autoregressive losses of different modalities to the same interval, thereby improving training stability and preventing imbalance across modalities.

\subsection{Once-For-All with Elastic Training}
\label{sec:elastic-training}

\begin{figure}[t]
  \centering
    \includegraphics[width=0.9\linewidth]{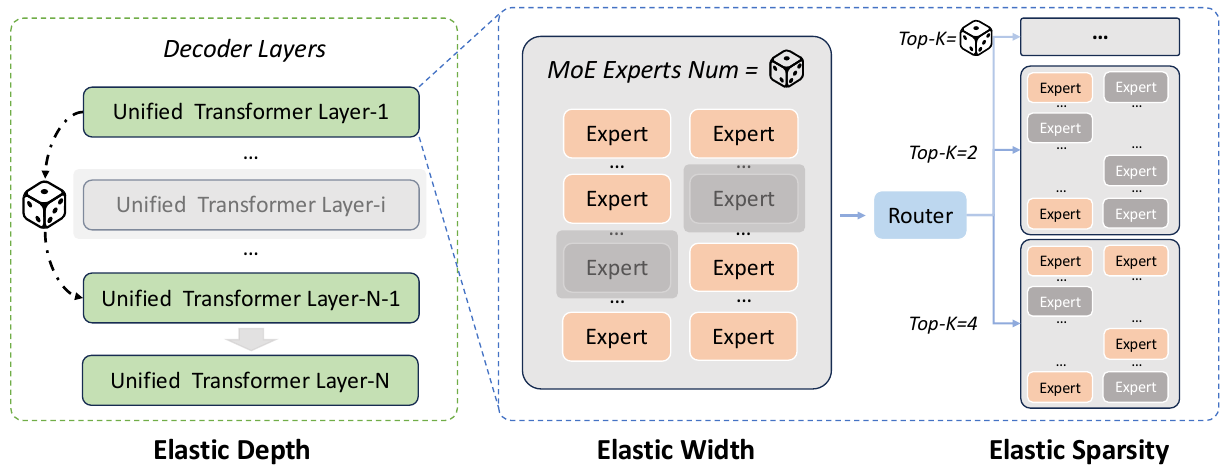} 
    \caption{Overview of the elastic training framework in \ourmodel. The framework supports elastic depth, width, and sparsity in a unified MoE architecture. \emph{Elastic Depth} randomly adapts the number of active layers, \emph{Elastic Width} varies the total number of experts in each MoE layer, and \emph{Elastic Sparsity} changes the top-$k$ routing per token. These mechanisms collectively enable flexible deployment under different compute, memory, and latency constraints without retraining.}
\label{fig:elastic_training}
\end{figure}

The scaling of modern models presents a fundamental challenge. Although models with trillions of parameters achieve remarkable performance across a wide range of tasks, their high computational and deployment costs limit the applicability in scenarios that demand flexibility and efficiency. 

Traditional approaches typically follow a ``train-then-compress'' pipeline, employing techniques such as pruning~\citep{sajjad2023effect, xia2023sheared, men2025shortgpt}, knowledge distillation~\citep{gu2023minillm, xu2024survey} or more efficient fusion-based variants~\citep{wang2023learning, chen2024lemon, chen2024dha} to produce smaller models. 
However, this paradigm still suffers from notable limitations. Model compression requires a dedicated pruning or distillation stage, which demands specialized infrastructure and incurs substantial computational overhead. 
Moreover, once a model is compressed, its architecture becomes fixed. Consequently, creating models of other sizes necessitates repeating the full compression process, thereby constraining deployment flexibility.

To address these issues, we propose a novel elastic training strategy and apply it to \ourmodel for the first time. Rather than compressing a pre-trained model post hoc, elastic training simultaneously optimizes a family of sub-networks during pre-training, so that a single large model to efficiently produce smaller, deployable variants on demand.
It extends the design philosophy of Once-For-All ~\citep{devvrit2023matformer,cai2024flextron,gu2025elastic} to pre-training, in which sub-network configurations of varying depth, width, and sparsity are trained together with the full-scale model.
As a result, \ourmodel can flexibly selects subsets of parameters to construct models at different scales, which reduces the computational and engineering overhead compared with traditional pruning or distillation methods.

The elastic training is shown in Figure~\ref{fig:elastic_training}, which introduces structural flexibility along three orthogonal dimensions:

\paragraph{Elastic Depth}
To support elastic depth, \ourmodel randomly varies the number of active transformer layers during training, enabling the extraction of sub-networks with different depths. 
Most of the time, the full-depth network is used to ensure all layers are well-optimized, while shallower sub-networks are occasionally sampled to foster resilience against layer removal.
Specifically, the full model is trained with a probability of $75\%$, while a reduced-depth sub-network is activated with a probability of $25\%$. 
Through this training scheme, intermediate representations are encouraged to remain informative even when some layers are bypassed, and the model supports flexible deployment across different depth configurations without requiring separate training.

\paragraph{Elastic Width}
Complementary to elastic depth, \ourmodel also supports elastic width by varying the total number of experts in each Mixture-of-Experts (MoE) layer.
Instead of always activating the full expert pool, the training process alternates between two modes.
With a probability of $80\%$, all experts participate in routing, preserving the full-width configuration. In the remaining $20\%$ of cases, routing is restricted to a randomly sampled subset of experts, leading to a narrower effective model width.
By exposing the model to both full and reduced expert configurations, the resulting model supports different capacity budgets with only partial experts, making it suitable for deployment in memory-constrained environments where hosting all experts is impractical.

\paragraph{Elastic Sparsity}
To improve inference efficiency without changing the deployed model size, elastic sparsity is introduced by varying the number of activated experts per token.
Similar to the elasticity of total number of experts, the default routing configuration is applied with a probability of $80\%$ during training. 
With a probability of $20\%$, the routing top-$k$ is randomly sampled from a predefined range, where $k$ is smaller than the standard configuration. 
In other words, the number of activated experts for each token is decreased.
Finally, the model is compatible with different compute budgets and exhibits improved robustness during latency-constrained inference.

By training an elastic super-network, \ourmodel is able to produce smaller models of varying configurations by selecting subsets of parameters along the layer number, total expert number, and activated expert number. Elasticity along the representation dimension (i.e., hidden size), as explored in recent work~\citep{chen2024mixture}, is orthogonal to our design and can be naturally incorporated as a future extension. These sub-networks can be instantiated on demand to meet different latency and memory constraints, serving as effective starting points for mid-training or fine-tuning.
Compared with training separate models from scratch or relying on post-hoc compression techniques, our elastic training strategy substantially lowers overall computational overhead and engineering complexity.
Detailed experimental results and ablation studies are provided in Sec~\ref{sec:elastic-discussion}.

\section{Post-Training}
After unified pre-training, we follow the same post-training pipeline as ERNIE 4.5~\citep{ernie45tech} to obtain the final \ourmodel, which includes two stages, supervised fine-tuning (SFT) and unified multimodal reinforcement learning (UM-RL). 
With curating comprehensive set of high-quality instruction pairs, SFT endows the model with fundamental instruction-following capability and strengthens its ability to think through long chains-of-thought. 
During UM-RL phase, we merge the training of various tasks such as reasoning, agent, and instruction following into a multi-stage RL pipeline, enabling balanced performance across diverse tasks and modalities. 
We further extended the unified verifier system to generate accurate and consistent reward signals for model responses in a wide range of multimodal scenarios, providing reliable supervision for unified multimodal RL training. 
In this section, we will discuss the key challenges of RL training and describe the solutions we propose.

The RL training of \ourmodel faces several challenges.
Firstly, RL training is computationally expensive, which is further amplified by the large scale of \ourmodel.
Secondly, the ultra-sparse MoE architecture exacerbates the training–inference discrepancy and undermines stability. % of RL training. 
Finally, compared to standalone RLVR tasks such as mathematical reasoning or code generation, training a model that simultaneously supports multiple scenarios and modalities introduces substantially higher complexity. To address these bottlenecks, we implement a suite of synergistic engineering and algorithmic optimizations that enable stable RL training for large-scale, ultra-sparse MoE models. In the sections that follow, we delineate the formidable challenges encountered in this endeavor and present our corresponding solutions.

\begin{figure}[t]
  \centering
    \includegraphics[width=0.9\linewidth]{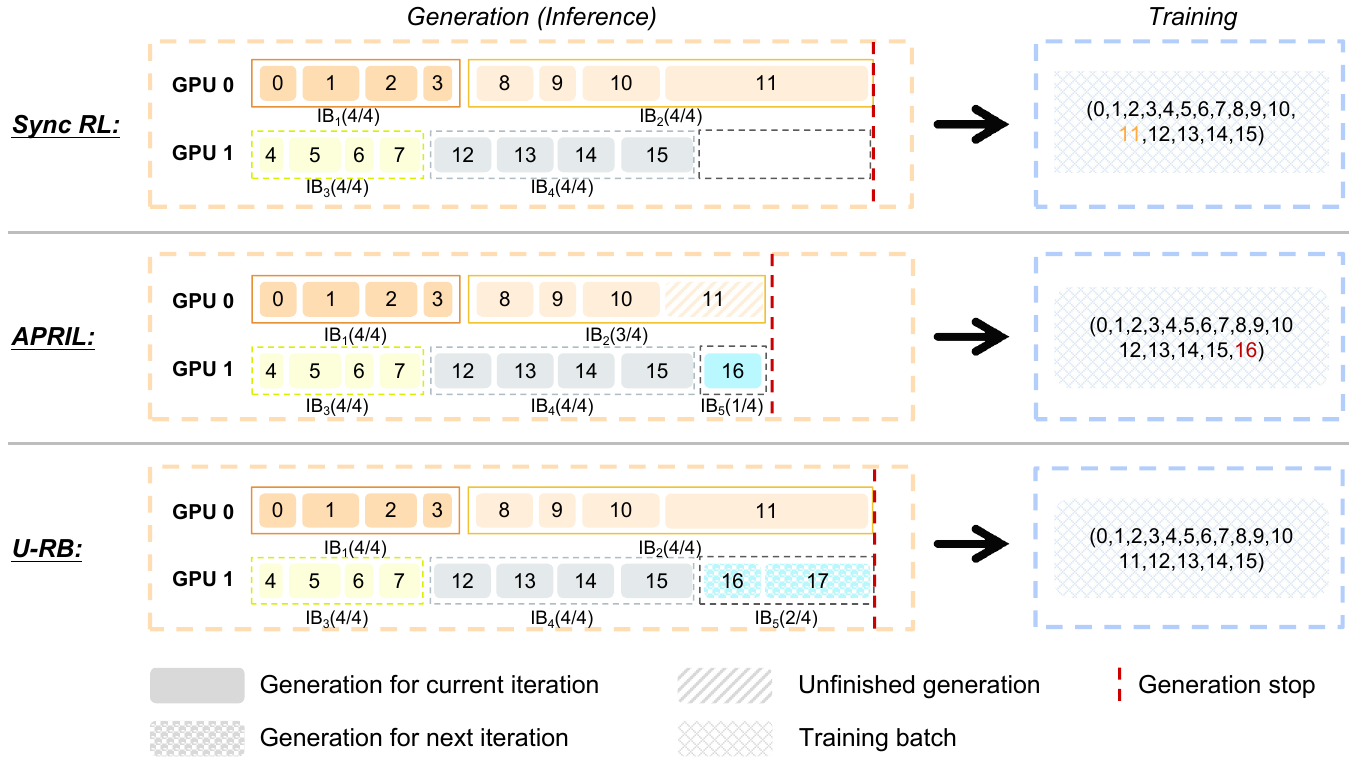} 
    \caption{
    Visualization of the Unbiased Replay Buffer (U-RB) in \ourmodel, in comparison with existing methods, where each query is assigned a unique index, and IB denotes the inference batch.
    In Sync RL, a long-tail query (e.g., index 11) blocks the entire batch, leaving GPUs idle and poorly utilized.
    APRIL stops generation once the target number of responses (16) is reached, which leads to a non-stationary data difficulty distribution.
    U-RB extends APRIL with a data-ordering constraint that prepares future batches while waiting for long-tail queries, preserving query order and mitigating inefficiency.}
    \label{fig:urb}
\end{figure}

\subsection{Enhancing Rollout Efficiency with Unbiased Replay Buffer}
\label{subsec:urb}
Rollout generation accounts for more than 90\% of the total training time in RL, and efficiency is often limited by the long-tail distribution of rollout response lengths. In that case, a small number of unusually long responses stall entire batches, leaving GPUs idle and underutilized.
Recent work, such as \text{APRIL}~\citep{zhou2025april}, seeks to mitigate long-tail inefficiency by over-provisioning rollout requests.
Generation is terminated once a target number of responses is collected, and incomplete responses are recycled for continuation in subsequent steps.
However, APRIL tends to update model parameters using trajectories with shorter reasoning steps, which usually correspond to easier queries.
In contrast, longer-horizon samples are deferred, leading to a non-stationary distribution of data difficulty.
Consequently, periodic shifts in data difficulty may hinder convergence and ultimately degrade model performance.  

\paragraph{U-RB: Unbiased Replay Buffer Generation} 
We introduce U-RB, an unbiased extension of APRIL that accelerates rollout generation in RL.
As illustrated in Figure~\ref{fig:urb}, U-RB introduces a data-ordering constraint, under which only the data group assigned to the current iteration at initialization is allowed to participate in subsequent training process. 
Specifically, U-RB builds two modules.
The first is a high-throughput inference pool, $\mathcal{P}_{infer}$, with capacity $\Omega_{RBS}=\Omega_{BS} * N$, where $\Omega_{BS}$ is the training batch size and $N$ is the buffer size, 
The second component is a training pool $\mathcal{P}_{train}$ with capacity $\Omega
_{BS}$, which collects completed trajectories for RL training.
At iteration $t$, the inference engine $\pi_{{infer};\theta_{t}}$ populates the inference pool by generating rollouts in parallel. 
Inference proceeds until the terminal state (i.e., $[EOS]$) is reached for the longest rollout belonging to the data group $\mathcal{D}_{t}$ assigned to iteration $t$. At this point, rollouts associated with $\mathcal{D}_{t}$ are moved from $\mathcal{P}_{infer}$ to the training pool $\mathcal{P}_{train}$, enabling the training engine $\pi_{train;\theta_{t}}$ to update model parameters.
These rollouts may include trajectories resumed from earlier inference runs.
By dynamically partitioning rollout generation, U-RB prevents computational idleness caused by individual long rollouts, while maintaining an unbiased data distribution.

\subsection{Stabilizing Training with Mitigated Entropy Collapse}
The phenomenon of rapid entropy collapse in a multimodal model is manifested as a  sharp increase or decrease in policy entropy during the early stages of RL. 
In multimodal decision-making tasks that integrate text, vision and audio information, such collapse gradually erodes the model’s ability to fuse information across modalities for flexible reasoning and reveals a pronounced modality bias.

Recent studies~\citep{cui2025entropy,wang2025beyond} attribute entropy collapse mainly to two factors. 
First, most contemporary RL frameworks rely on separate engines for training and inference, which introduces inconsistencies in numerical computation, and ultimately destabilizes policy optimization. 
The problem becomes more severe for MoE models, where dynamic routing further amplifies the numerical mismatch problem. 
Second, the policy model often overfits easy queries in the early stage of training.
Such behavior accelerates entropy collapse and limits the model’s ability to discover alternative reasoning paths.
To address these issues, we introduce Multi-granularity Importance Sampling Clipping (MISC) and Well-learned Positive Sample Mask (WPSM) to stabilize RL training at scale.

\begin{figure}[t]
  \centering
  \begin{minipage}[t]{0.49\linewidth}
    \centering
    \includegraphics[width=0.9\linewidth]{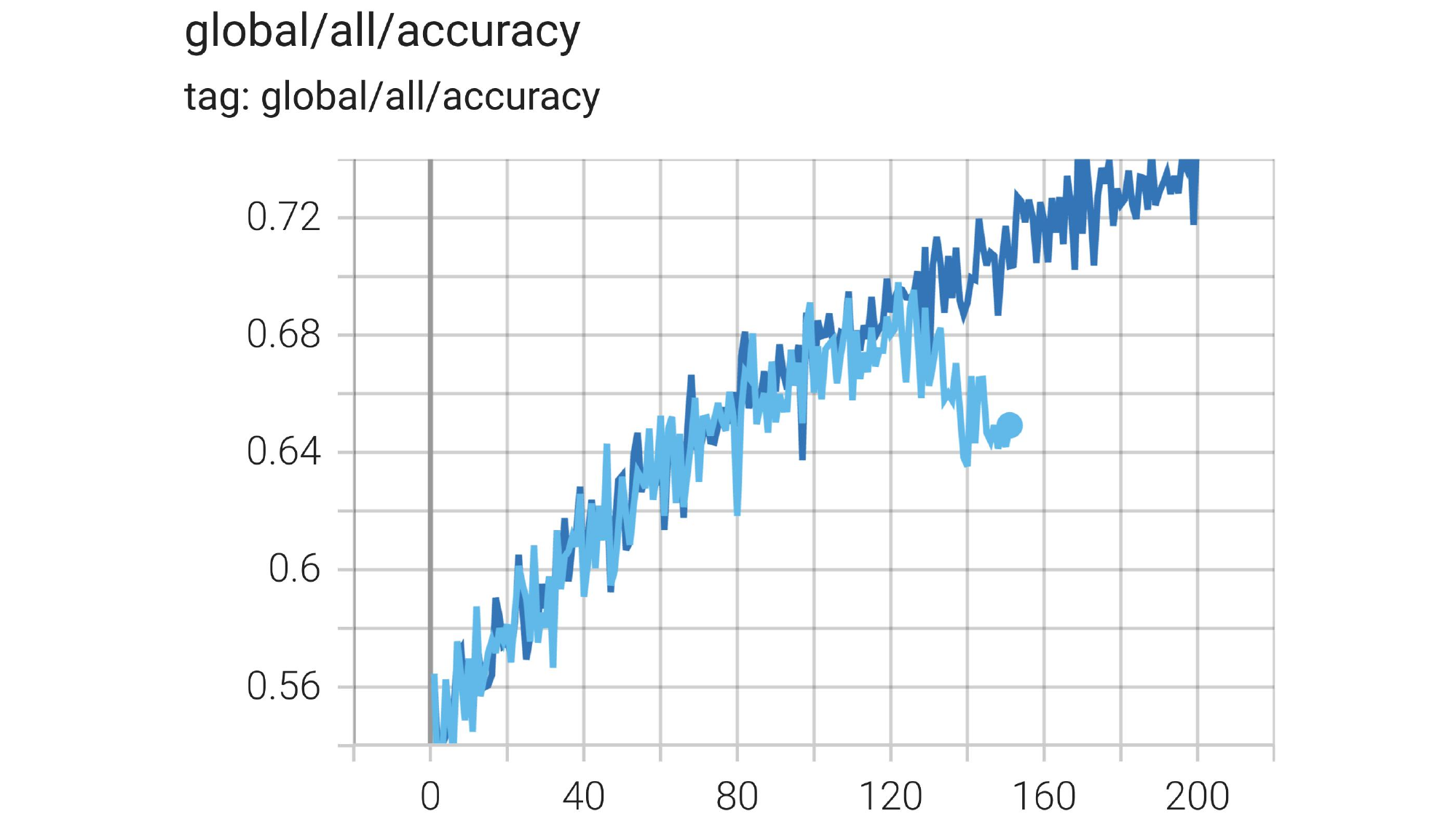} 
  \end{minipage}
  \begin{minipage}[t]{0.49\linewidth}
    \centering
    \includegraphics[width=0.9\linewidth]{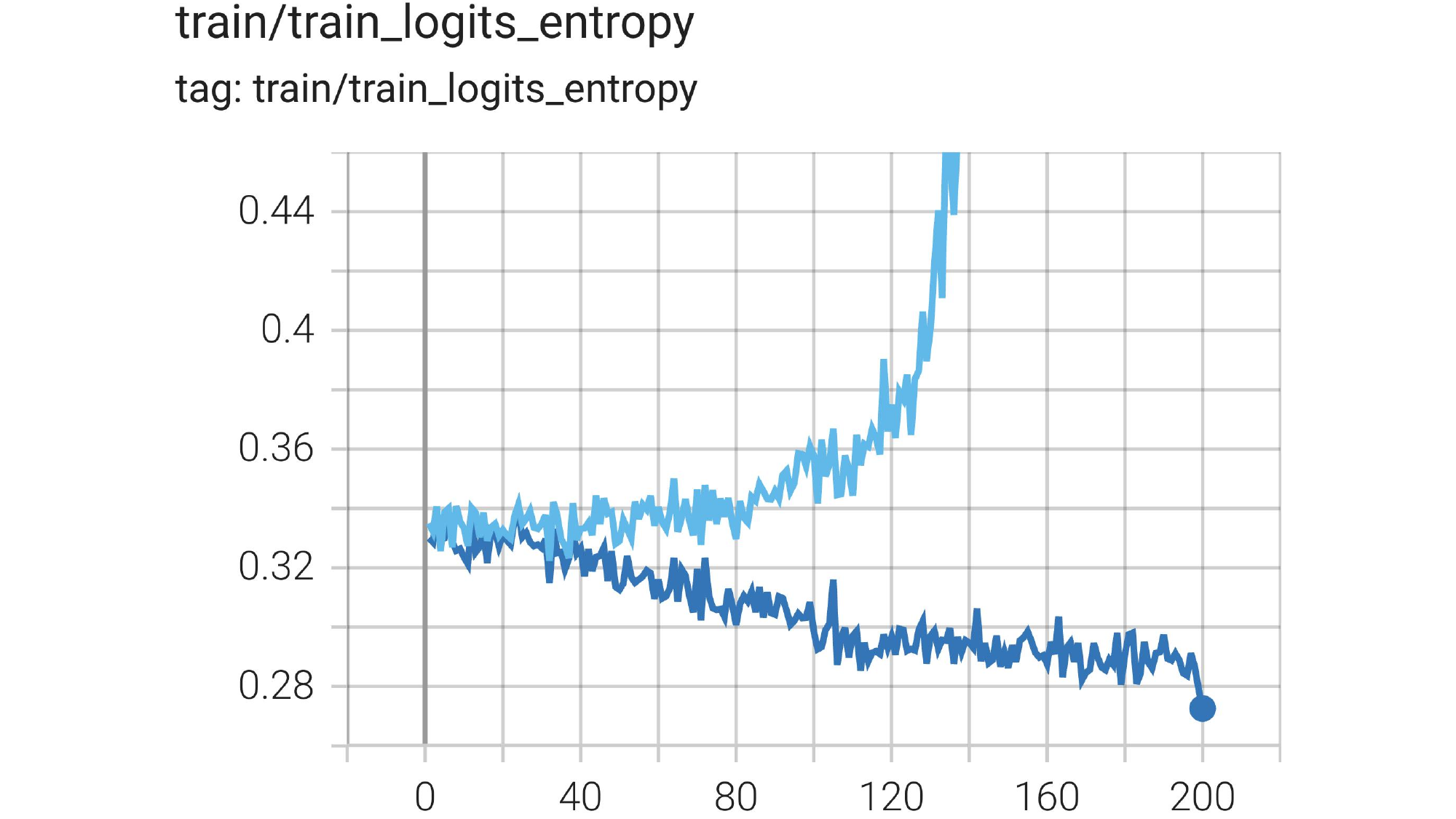} 
  \end{minipage}
  \caption{Training dynamics of applying $\mathfrak{J}_{IcePop}^{Mixed}$ (dark-blue) and $\mathfrak{J}_{IcePop}^{GSPO}$ (light-blue) to conduct RL training on \ourmodel. By using Multi-granularity Importance Sampling Clipping (MISC), we avoid the entropy collapse during early stage and achieve stable RL training.}
   \label{fig:mis}
\end{figure}

\paragraph{MISC: Multi-granularity Importance Sampling Clipping}
IcePop \citep{team2025every} suppresses training-inference mismatch through double-sided masking calibration on GRPO~\citep{deepseekr1}:
\begin{equation}
\begin{aligned}
    \mathfrak{J}_{IcePop}^{GRPO}(\theta) &= \mathbb{E}_{x \sim D,\{y_i\}_{i=1}^G \sim \pi_{infer}(\cdot|x;\theta_{old})} \\
    & ~~~~ \left[\frac{1}{G}\sum_{i=1}^{G}\frac{1}{|y_i|}\sum_{j=1}^{|y_i|}[\mathfrak{M}(\frac{\pi_{train}(y_{i,j}|x,y_{i,<j};\theta_{old})}{\pi_{infer}(y_{i,j}|x,y_{i,<j};\theta_{old})};\alpha,\beta) \cdot min(r_{i,j}\hat{A_{i,j}},~clip(r_{i,j}, 1-\epsilon, 1+\epsilon)\hat{A_{i,j}}\right] \\
    & ~~~~~ r_{i,j} = \frac{\pi_{train}(y_{i,j}|x,y_{i,<j};\theta)}{\pi_{train}(y_{i,j}|x,y_{i,<j};\theta_{old})} \\
    & ~~~~~\mathfrak{M}(k) = 
    \begin{cases}
        k & if ~k \in [\alpha, \beta], \\
        0 & otherwise
    \end{cases}
\end{aligned}
\end{equation}
where $\alpha, \beta$ controls the lower and upper limits. We apply this technology to GSPO~\citep{zheng2025group}:
\begin{equation}
\begin{aligned}
    \mathfrak{J}_{IcePop}^{GSPO}(\theta) &= \mathbb{E}_{x \sim D,\{y_i\}_{i=1}^G \sim \pi_{infer}(\cdot|x;\theta_{old})} \\
    & ~~~~ \left[\frac{1}{G}\sum_{i=1}^{G}[\mathfrak{M}(\left(\frac{\pi_{train(y_i|x;\theta_{old})}}{\pi_{infer(y_i|x;\theta_{old})}}\right)^{\frac{1}{|y_i|}};\alpha,\beta) \cdot min(s_i(\theta)\hat{A}_i,~clip(s_i(\theta), 1-\epsilon, 1+\epsilon)\hat{A_i}\right] \\
    & ~~~~~ s_i(\theta) = \left(\frac{\pi_{train}(y_i|x;\theta)}{\pi_{train}(y_i|x;\theta_{old})} \right)^{\frac{1}{|y_i|}} = exp\left(\frac{1}{|y_i|}\sum_{j=1}^{|y_i|}log\frac{\pi_{train}(y_{i,j}|x,y_{i,<j};\theta)}{\pi_{train}(y_{i,j}|x,y_{i,<j});\theta_{old})}\right)\\
    & ~~~~~\mathfrak{M}(k) = 
    \begin{cases}
        k & if ~k \in [\alpha, \beta], \\
        0 & otherwise
    \end{cases}
\end{aligned}
\end{equation}
However, our experiments show that directly applying $\mathfrak{J}_{IcePop}^{GSPO}$ to RL training on \ourmodel leads to rapid entropy collapse, as illustrated by the light-blue line in Figure~\ref{fig:mis}. 
The phenomenon is caused by sequence-level truncated importance sampling, which prunes a large number of low-entropy responses due to the training–inference mismatch.

To address this issue, we revise $\mathfrak{J}_{IcePop}^{GSPO}$ to $\mathfrak{J}_{IcePop}^{Mixed}$:
\begin{equation}
\begin{aligned}
    \mathfrak{J}_{IcePop}^{Mixed}(\theta) &= \mathbb{E}_{x \sim D,\{y_i\}_{i=1}^G \sim \pi_{infer}(\cdot|x;\theta_{old})} \\
    & ~~~~ \left[\frac{1}{G}\sum_{i=1}^{G}[\mathfrak{M}_{j \in [1, |y_i|]}(\frac{\pi_{train}(y_{i,j}|x,y_{i,<j};\theta_{old})}{\pi_{infer}(y_{i,j}|x,y_{i,<j};\theta_{old})};\alpha,\beta) \cdot min(s_i(\theta)\hat{A}_i,~clip(s_i(\theta), 1-\epsilon, 1+\epsilon)\hat{A_i}\right] \\
    & ~~~~~ s_i(\theta) = \left(\frac{\pi_{train}(y_i|x;\theta)}{\pi_{train}(y_i|x;\theta_{old})} \right)^{\frac{1}{|y_i|}} = exp\left(\frac{1}{|y_i|}\sum_{j=1}^{|y_i|}log\frac{\pi_{train}(y_{i,j}|x,y_{i,<j};\theta)}{\pi_{train}(y_{i,j}|x,y_{i,<j});\theta_{old})}\right)\\
    & ~~~~~\mathfrak{M}(k) = 
    \begin{cases}
        k & if ~k \in [\alpha, \beta], \\
        0 & otherwise
    \end{cases}
\end{aligned}
\end{equation}

By modulating the trust region according to the modality sensitivity, we achieve a more balanced exploration–exploitation trade-off.
The mechanism avoids premature convergence to a ``safe'' yet suboptimal strategy in complex multi-scenario settings, and preserves flexibility across various inputs.

\paragraph{WPSM: Well-learned Positive Sample Mask}
We introduce a sample mask strategy to prevent the model from over-optimizing on already mastered queries, in which the proficiency is tracked by maintaining a success-rate for each query. 
For a given query $x$ with a rollout group $\mathcal{Y}^x=\{y^x_1,y^x_2,...,y^x_G\}$, where $G$ is the group size,
if the average accuracy $acc^{x}_{t}$ in iteration $t$ exceeds a threshold $\tau$, we flag the rollout $y^x_i$ in $\mathcal{Y}^x$ as a ``well-learned'' response when its policy entropy $\mathcal{H}_{y^x_i}(\pi_\theta)$ falls below a stability bound $\eta$. 
During training, the ``well-learned'' responses are masked as follows:
\begin{equation}
\begin{aligned}
    \mathfrak{J}(\theta) &= \mathbb{E}_{x \sim D,\{y_i\}_{i=1}^G \sim \pi_{\theta_{old}}(\cdot|x)} 
    \left[\frac{1}{G}\sum_{i=1}^{G}[1-\mathbb{M}_{mask}^i]min(s_{i}(\theta))\hat{A_i},~clip(s_i(\theta), 1-\epsilon, 1+\epsilon)\hat{A_i}\right] \\
    & ~~~~~\mathbb{M}_{mask}^{i} = 
    \begin{cases}
        \alpha & \mathcal{H}_{y^x_i}(\pi_\theta) < \eta ~~ and ~~ acc^{x}_{t} > \tau \\
        0 & otherwise
    \end{cases}
\end{aligned}
\end{equation}
%  defined in~\cite{zheng2025group},
where $s_i(\theta)$ denotes the importance ratio and $\alpha \in [0,1]$ controls the degree of supplementary learning applied to ``well-learned'' responses. 
Under this design, the gradient budget is shifted toward harder samples, such as those with sparse rewards or diverse reasoning paths. 
By masking redundant positive signals, WPSM alleviates the entropy collapse problem caused by over-fitting to easy queries, and encourages the model to improve the performances of challenging, low-performing tasks.

\begin{figure}[t]
  \centering
    \includegraphics[width=0.8\linewidth]{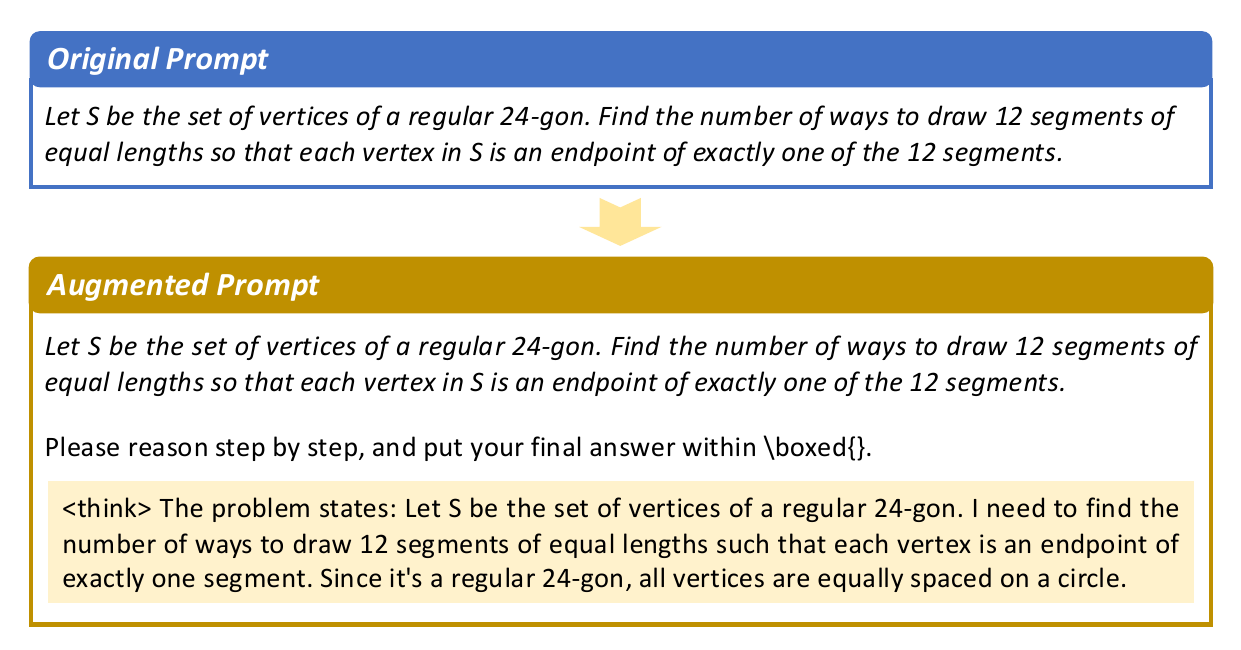} 
    \caption{Overview of the Adaptive Hint-based Reinforcement Learning (AHRL) in \ourmodel, which introduces think skeletons to guide hard queries and mitigate sparse rewards.}
    \label{fig:ahrl}
\end{figure}

\subsection{Boosting Sample Efficiency with Hint-based Learning}
Recent studies~\citep{yue2025does,liu2025prorl,zhao2025echo} indicate that although state-of-the-art RL methods, such as GRPO~\citep{deepseekr1} and DAPO~\citep{yu2025dapo}, enhance the pass@1 metric by reinforcing high-reward completions, they exhibit clear limitations on challenging tasks where the base model performs poorly.
Specifically, when all rollouts receive zero reward, the GRPO framework fails to provide effective gradient signals for policy optimization.
In such cases, RL training on hard queries tends to progress much more slowly because sparse rewards and limited sample efficiency impede learning.
To address this challenge, we propose Adaptive Hint-based Reinforcement Learning (AHRL), a method that mitigates the issues of sparse rewords on hard queries. 
As can be seen in Figure~\ref{fig:ahrl}, AHRL introduces partial hints that decompose the complex problem into intermediate steps and gradually increase the performance of trained models.
The mechanism is described in detail below.

\paragraph{AHRL: Adaptive Hint-based Reinforcement Learning}
Unlike approaches that modify reward functions or optimization algorithms, AHRL is designed to inject partial think sketches into queries during RL training. By decomposing problems into intermediate steps, AHRL increases the propensity of the base model to generate correct responses and improves sample efficiency.
As a result, the model is driven to master the hardest problems, which accelerates the RL training process.

For a given query $x$ with a response consisting of a thinking trajectory and a final solution, denoted as $y = (think, solution)$, AHRL augments $x$ into $\tilde{x}^{(p)}$ by attaching the first $p_{hint}$ tokens of the think to the original query.
$p_{hint}$ denotes the fraction of revealed thinking, allowing fine-grained control over query difficulty. 
Specifically, The probability $p_{hint}$ follows an annealing schedule:
\begin{equation}
    p_{hint}(x^{t}) = p_{initial} \cdot \exp(-\gamma \cdot t \cdot pass_{initial}^{x})
\end{equation}
where $t$ is the training iteration, $\gamma$ is the decay rate, and $pass_{initial}^{x}$ is the pass@k score of query $x$ evaluated on the SFT model.
As training progresses and model performance improves, the fraction of revealed hints is gradually reduced, transitioning the model to full self-exploration.
The mechanism provides necessary ``scaffolding'' to bridge the gap between initial exploration and successful task completion. 
It prevents training from stalling in complex tasks where valid reasoning paths are statistically rare, and ensures consistent performance improvements across modalities.

\section{Infrastructures}
\label{sec:infra}

The training of \ourmodel is built upon PaddlePaddle~\citep{paddlepaddle}. Based on the infrastructure of ERNIE~4.5~\citep{ernie45tech}, we further address the unique challenges introduced by native multimodal training, ultra-sparse MoE models, and large-scale RL pipelines.

\paragraph{Tremendous Memory Pressure and Communication Overhead}
The ultra-sparse MoE architecture of \ourmodel poses two major challenges  for efficient training: ultra-sparse expert activation leads to heavy inter-node communication, and large-scale expert parameters impose significant memory pressure.
To mitigate these challenges, we propose a hybrid parallel strategy tailored for the MoE architecture, integrated with fine-grained memory control.

\paragraph{Multimodal Training with Multiple Tokenizers}
As a unified model supporting text, vision, and audio, \ourmodel relies on multiple modality-specific tokenizer models.
Their computational characteristics differ significantly from those of the MoE backbone, making it challenging to apply traditional end-to-end optimization strategy across all parts.
To resolve this mismatch, we adopt a decoupled architecture that physically separates tokenizers from the MoE backbone and deploys them on different GPU nodes, allowing each component to use its most suitable parallelization strategy.

\paragraph{Flexible Attention Patterns across Modalities}
\ourmodel processes inputs from multiple modalities with heterogeneous attention patterns.
Visual inputs typically require bidirectional attention, whereas text and audio inputs rely on unidirectional attention.
Although existing solutions such as FlexAttention~\citep{dong2024flex} support flexible attention, they are less efficient when attention mask patterns vary across samples within the same batch.
We therefore adopt the self-developed FlashMask~\citep{flashmask} to accelerate attention mask computation here.

\paragraph{Scalable RL under Throughput and Consistent Constraints}
Reinforcement learning for a trillion-parameter model requires coordinated execution across training, inference, and environment interaction.
Such heterogeneous workloads pose challenges in maintaining numerical consistency between training and rollout, mitigating data distribution bias in asynchronous pipelines, and maximizing the utilization of diverse hardware resources.
We therefore design a scalable and disaggregated RL infrastructure to resolve these bottlenecks while preserving strict computational determinism.

\subsection{Hybrid Parallelism for Training at Scale}
To address the severe memory pressure and communication overhead introduced by the ultra-sparse MoE architecture, we develop a distributed parallel strategy that adapts to varying training resources.
The final configuration combines 4-way tensor parallelism~\citep{megatronlm}, 12-way pipeline parallelism~\citep{gpipe} with virtual stages, 64-way expert parallelism~\citep{gshard}, ZeRO-1 data parallelism~\citep{zero}, and context parallelism~\citep{ringattention} for long context training. 
We also use DeepEP~\citep{deepep} to enable efficient inter-node communication, while virtual pipeline parallelism is employed to minimize pipeline bubbles.
To ensure final performance, \ourmodel utilize a no-token-dropping strategy throughout training, and out-of-memory (OOM) issues sometimes occur due to unbalanced expert routing, especially in the initial stage of training.
To mitigate these issues and stabilize large-scale MoE training, we develop a set of techniques as follows.

To ensure memory sufficiency, we implement the following strategies:
\begin{itemize}
    \item \textbf{FP8 Mixed-Precision Training}. 
    Following the practice of ERNIE~4.5~\citep{ernie45tech}, we adopt FP8 mixed-precision training and store activation tensors in FP8 format, which effectively reduces peak memory consumption during training.
    \item \textbf{Dynamic Adaptive Offloading of Activation Memory}. 
    During forward propagation, all activation tensors retained for backward computation are tracked.
    We extend the memory allocator to enable adaptive offloading of selected activation tensors when an OOM event is encountered.
    No offloading is triggered when sufficient memory is available.
    This technology ensures sufficient total memory for training with minimal performance overhead.
\end{itemize}

To reduce memory fragmentation, we implement the following strategies:
\begin{itemize}
    \item \textbf{Sub-batch Computations}. 
    We decompose large memory allocation requests into a series of smaller requests with sub-batch computation, which reduces the probability of OOM caused by memory fragmentation.
    \item \textbf{Automatic Memory Defragmentation}. 
    Based on CUDA virtual memory management (VMM), we develop a memory allocation allocator capable of automatic memory defragmentation, ensuring successful memory allocation even under extreme conditions.
\end{itemize}

Through the aforementioned techniques, we ensure the feasibility and reliability of pre-training \ourmodel in memory-constrained scenarios.

\subsection{Disaggregation Architecture for Multimodal Training}
\ourmodel integrates heterogeneous multimodal inputs, where each modality is handled by a dedicated tokenizer that converts raw signals into token sequences.
Both sequence lengths and computational costs vary substantially across modalities and even among samples within the same modality.
Deploying all tokenizers together with backbone on homogeneous hardware would induce significant load imbalance, which in turn degrades overall training efficiency.

To solve this problem, we design a tokenizer-backbone disaggregation architecture. 
We decouple the tokenizers from the backbone by deploying them as independent, horizontally scalable services on dedicated compute nodes, under a data-parallel configuration.
During training, the backbone interacts with these tokenizer services via remote calls to retrieve encoded representations. 
The architectural separation allows each component to adopt parallelization strategies suited to its own workload, improving scalability and efficiency in distributed multimodal training.

\subsection{FlashMask for Flexible Multimodal Attention}
In \ourmodel, attention patterns vary across modalities and may even differ across samples.
Text modeling typically relies on causal attention, while visual features often employ bidirectional attention.
That is, for visual samples, attention follows a globally causal structure while allowing local bidirectional interactions to capture spatial dependencies.
Efficiently supporting such heterogeneous masking patterns is therefore essential for unified multimodal training.

To meet these requirements, we employ FlashMask~\citep{flashmask}, which not only meets the flexible and diverse attention masking needs of \ourmodel, but also significantly accelerates the computational efficiency of attention mask operations.
In practice, FlashMask achieves up to a 200\% speedup over FlexAttention~\citep{dong2024flex} at the operator-level, and delivers more than 20\% end-to-end training acceleration. In addition, we integrate FlashMask with context parallelism~\citep{ringattention} at the kernel level, achieving an 80\% performance improvement compared to the Megatron-LM solution.

\subsection{Scalable and Disaggregated RL Infrastructure}

Scaling Reinforcement Learning (RL) to unified multimodal models with trillions of parameters presents unique challenges regarding computational consistency, data distribution bias, and heterogeneous resource utilization.
To address these bottlenecks, we introduce the \ourmodel RL Infrastructure, a disaggregated system designed to orchestrate large-scale asynchronous training. By prioritizing high-throughput execution and computation determinism, our system ensures stable and efficient RL training.

The key architectural components are summarized as follows:

\begin{itemize}
    \item \textbf{Disaggregated Control Plane for Asynchronous RL.}
    We introduce a fully disaggregated control plane built around a centralized RL controller to maximize system throughput, which coordinates training, inference, environment interaction, and reward evaluation in an asynchronous manner.
    Logical decoupling across these subsystems enables flexible scaling and efficient pipeline management, forming the foundation for large-scale asynchronous multimodal RL.

    \item \textbf{Unified FP8 Stack for Consistent Training and Inference.}
    Precision divergence is a common issue in low-bit RL training. To alleviate this situation, we build a unified FP8 execution engine.
    By employing identical high-performance operators across both training and inference (rollout) stages and integrating the Rollout Router Replay~\citep{ma2025stabilizing} strategy, the engine minimizes numerical mismatch and ensures stable convergence under low-precision settings.

    \item \textbf{Replay Buffer for Sequence-Length Bias Mitigation.}
    Asynchronous rollout in RL may introduce sequence-length bias, where shorter responses enter training earlier and distort the data distribution.
    We design an unbiased replay buffer, in collaboration with the algorithm described in Section~\ref{subsec:urb}, that preserves the original data order, ensuring consistent data arrival and mitigating bias caused by asynchronous completion.

    \item \textbf{Heterogeneous Resource Optimization with Elastic CPU Pooling.}
    To address the under utilization of CPU resources commonly observed in GPU-dominated AI clusters, we implement an elastic CPU pooling strategy.
    The elastic mechanism isolates and virtualizes idle CPU capacity from the cluster to power logic-intensive tasks such as intensive RL environment interactions and result verification.
    It effectively amplifies the computational resources available for environment rollouts, enabling massive-scale parallel simulation. Consequently, it reduces the wall-clock time of training iterations while significantly improving the total cost of ownership (TCO) efficiency of the underlying hardware.
\end{itemize}

\section{Evaluations}
\label{sec:evaluations}

We conduct systematic evaluations of \ourmodel against state-of-the-art models across a wide range of text (Sec.~\ref{sec:eval-text}), vision (Sec.~\ref{sec:eval-visual}), and audio benchmarks (Sec.~\ref{sec:eval-audio}) with internal evaluation framework, ERNIE-Eval\footnote{For code-related tasks, evaluations are conducted with LiveCodeBench~\citep{livecodebench} and SandboxFusion~\citep{sandboxfusion} to ensure reliable and execution-based assessment.}.
Next, we further provide an in-depth analysis of our two key design choices, namely \emph{modality-agnostic expert routing} (Sec.~\ref{sec:moe-routing-discussion}) and \emph{elastic training} (Sec.~\ref{sec:elastic-discussion}).

\subsection{Evaluation on Language Benchmarks}
\label{sec:eval-text}

To comprehensively assess the text-centric capabilities learned during large-scale pre-training and post-training, we evaluate \ourmodel on a diverse set of benchmarks covering factual knowledge, reasoning, mathematical problem solving, coding, multilingual understanding, instruction following and agent-oriented tasks. 
These benchmarks are selected to encompass both core language modeling abilities and advanced reasoning and decision-making skills. This enables a systematic examination of how the unified architecture and training strategies translate into downstream text performance.

\paragraph{Benchmarks}
\begin{itemize}
    \item \textbf{Knowledge:}   PreciseWikiQA~\citep{precisewikiqa}, PopQA~\citep{popqa}, HotPotQA~\citep{yang2018hotpotqa},
    ChineseSimpleQA~\citep{chinese-simpleqa},
    SimpleQA~\citep{simpleqa}.
    \item \textbf{General:} MMLU-Pro~\citep{mmlu-pro}, MMCU~\citep{mmcu}, AGIEval~\citep{agieval}, MMLU~\citep{mmlu}, MMLU-Redux~\citep{mmlu-redux}, C-Eval~\citep{ceval}, CMMLU~\citep{cmmlu}, BBH~\citep{bbh}, WinoGrande~\citep{winogrande}, Humanity's Last Exam (HLE)~\citep{hle}.
    \item \textbf{STEM:} MATH (CoT)~\citep{math}, GPQA-Diamond~\citep{gpqa}, AIME 2025~\citep{aime}, HMMT 2025~\citep{hmmt2025}.
    \item \textbf{Coding:} LiveCodeBench v6 (2408 to 2505)~\citep{livecodebench}, HumanEval+~\citep{humaneval}, MBPP+~\citep{mbpp}, CRUXEval~\citep{gu2024cruxeval}.
    \item \textbf{Multilingual:} MMMLU~\citep{mmlu}, INCLUDE~\citep{romanou2024include}.
    \item \textbf{Reasoning:} ZebraLogic~\citep{zebralogic}, BBEH~\citep{bbeh}.
    \item \textbf{Instruct Following:} IFEval~\citep{ifeval}, Multichallenge~\citep{multichallenge}, Multi-IF~\citep{multiif}.
    \item \textbf{Agent:} TAU2-Bench~\citep{tau2bench}, ACEBench~\citep{chen2025acebench}, BFCL v4~\citep{patil2025bfcl}, BrowseComp~\citep{wei2025browsecomp}, SpreadSheetBench~\citep{ma2024spreadsheetbench}.
\end{itemize}

\paragraph{Evaluation of Pre-trained Models.}

\begin{table}[t]
    \centering
    \footnotesize
    \begin{tabular}{llccc}
        \toprule
         \textbf{Category} & \textbf{Benchmark} & \textbf{DS V3.2-Exp-Base} & \textbf{Kimi K2-Base} & \textbf{\basemodel} \\
        \midrule
        \multirow{4}{*}{Knowledge} 
        & PreciseWikiQA{\tiny 10-shot} & 52.60 & 61.66 & \textbf{74.48} \\
        & PopQA{\tiny 10-shot} & 48.73 & 51.74 & \textbf{65.24} \\
        & HotPotQA{\tiny 10-shot} & 53.11 & 57.07 & \textbf{67.08} \\
        & ChineseSimpleQA{\tiny 10-shot} & 74.19 & 78.29 & \textbf{90.09} \\
        \midrule
        \multirow{9}{*}{General} 
        & MMLU-Pro{\tiny 5-shot} & 68.27 & 67.19 & \textbf{75.58} \\
        & MMCU{\tiny 5-shot} & 91.16 & 90.63 & \textbf{93.92} \\
        & AGIEval{\tiny 5-shot} & 78.20 & 76.43 & \textbf{80.22} \\
        & MMLU{\tiny 5-shot} & 88.60 & 88.40 & \textbf{90.58} \\
        & MMLU-Redux{\tiny 5-shot} & 90.50 & 89.88 & \textbf{92.19} \\
        & C-Eval{\tiny 5-shot} & 90.70 & \textbf{92.49} & 91.98 \\
        & CMMLU{\tiny 5-shot} & 88.43 & \textbf{90.61} & 89.69 \\
        & BBH{\tiny 5-shot, w.o.-cot (Direct)} & 73.50 & 70.07 & \textbf{75.69} \\
        & WinoGrande{\tiny 5-shot} & 88.87 & 87.61 & \textbf{92.66} \\
        \midrule
        \multirow{2}{*}{STEM} 
        & MATH (CoT){\tiny 8-shot} & 65.70 & 65.90 & \textbf{73.89} \\
        & GPQA-Diamond{\tiny 5-shot} & 53.01 & 48.10 & \textbf{57.30} \\
        \midrule
        \multirow{5}{*}{Coding} 
        & LiveCodeBench v6{\tiny 1-shot} & 24.90 & 26.30 & \textbf{31.94} \\
        & HumanEval+{\tiny 0-shot} & 53.99 & 70.73 & \textbf{80.86} \\
        & MBPP+{\tiny 0-shot} & 78.77 & \textbf{79.36} & 79.10 \\
        & CRUXEval-I{\tiny 1-shot} & 59.54 & 73.64 & \textbf{79.75} \\
        & CRUXEval-O{\tiny 1-shot} & 71.28 & 81.61 & \textbf{84.01} \\
        \midrule
        \multirow{2}{*}{Multilingual} 
        & MMMLU{\tiny 5-shot} & 70.99 & 61.50 & \textbf{78.94} \\
        & INCLUDE{\tiny 5-shot} & 77.45 & 72.29 & \textbf{77.81} \\
        \bottomrule
    \end{tabular}%
    \caption{Comparison of pre-trained models on various language benchmarks. The best performance in each row is highlighted in \textbf{bold}.}
    \label{tab:pretrain-eval}
\end{table}

Table~\ref{tab:pretrain-eval} summarizes the pre-training results of \ourmodel in comparison with strong open-source baselines on a diverse set of text benchmarks.
Across these benchmarks, \ourmodel exhibits consistently strong and well-balanced performance in knowledge, reasoning, mathematics, coding, and multilingual tasks:

On knowledge-intensive benchmarks, \basemodel demonstrates clear and substantial advantages over DeepSeek V3.2-Exp-Base (DS V3.2-Exp-Base) and Kimi K2-Base, particularly on both English and Chinese question answering tasks.
The large margins observed on these datasets indicate that large-scale unified pre-training effectively consolidates factual knowledge and supports robust retrieval-style reasoning in multilingual settings.

On general reasoning and exam-style benchmarks, \basemodel achieves the best results on a wide range of challenging evaluations, including MMLU-Pro, MMLU, MMCU, AGIEval, MMLU-Redux, and BBH. Notably, the pronounced gains on harder benchmarks such as MMLU-Pro suggest improved reasoning depth and robustness, which can be attributed to the shared backbone and modality-agnostic MoE routing that encourage effective expert specialization.

On STEM tasks, \basemodel consistently outperforms strong baselines on MATH (CoT) and GPQA-Diamond, demonstrating robust multi-step reasoning and solution consistency. These improvements reflect enhanced long-horizon dependency modeling enabled by the deep unified architecture, together with stable optimization under elastic pre-training.

In the coding domain, \basemodel attains state-of-the-art performance on LiveCodeBench v6 and CRUXEval, while remaining competitive on MBPP+, indicating strong generalization of algorithmic and procedural reasoning learned during large-scale pre-training.

On multilingual benchmarks, \basemodel significantly surpasses baselines on MMMLU and INCLUDE, validating the effectiveness of unified tokenization and shared expert routing in learning robust multilingual representations.

The results demonstrate that the proposed unified architecture and elastic pre-training strategy jointly yield strong generalization across diverse text-centric tasks, providing a solid and versatile foundation for subsequent post-training and deployment.

\begin{table*}[t]
\centering
\footnotesize
\setlength{\tabcolsep}{2.1pt}
\resizebox{\textwidth}{!}{%
\begin{tabular}{llccccc}
\toprule
\textbf{Category} & \textbf{Benchmark} & \textbf{DS V3.2-Thinking} & \textbf{Gemini 2.5-Pro} & \textbf{GPT-5 (High)} & \textbf{Gemini 3-Pro} & \textbf{\ourmodel} \\
\midrule
\multirow{2}{*}{Knowledge} 
 & SimpleQA & 28.02 & 54.00 & 51.30 & 69.33 & \textbf{74.01} \\
 & ChineseSimpleQA & 72.37 & 76.50 & 75.10 & 84.08 & \textbf{86.03} \\
\midrule
\multirow{2}{*}{General} 
 & MMLU-Pro & 85.00 & 86.20 & \textbf{87.10} & 86.88 & 83.80 \\
 & HLE & 25.10 & 21.60 & 24.80 & \textbf{37.50} & 25.81 \\
\midrule
\multirow{3}{*}{STEM} 
 & GPQA-Diamond & 82.40 & 86.40 & 85.70 & \textbf{91.90} & 86.36 \\
 & AIME 2025 & 93.10 & 88.00 & 94.60 & \textbf{95.00} & 89.06 \\
 & HMMT 2025 & 86.67 & 81.20 & 93.30 & \textbf{93.33} & 79.58 \\
\midrule
\multirow{3}{*}{Coding} 
 & LiveCodeBench v6 & 81.06 & 72.90 & 81.70 & \textbf{86.34} & 76.21 \\
 & HumanEval+ & 90.80 & 94.50 & 92.70 & \textbf{95.12} & 94.48 \\
 & MBPP+ & 81.48 & 73.80 & 83.10 & \textbf{86.21} & 82.54 \\
\midrule
\multirow{2}{*}{Reasoning} 
 & ZebraLogic & 97.60 & 92.90 & \textbf{98.80} & 95.50 & 96.50 \\
 & BBEH & 67.04 & 68.80 & 69.00 & \textbf{78.80} & 66.63 \\
\midrule
\multirow{3}{*}{\shortstack[l]{Instruction\\Following}} 
 & IFEval & 91.87 & 89.50 & \textbf{94.10} & 92.24 & 93.35 \\
 & MultiChallenge & 42.43 & 51.50 & 58.30 & 62.50 & \textbf{65.98} \\
 & Multi-IF & 71.17 & 76.10 & 70.00 & 81.15 & \textbf{85.56} \\
\midrule
\multirow{6}{*}{Agent} 
 & TAU2-Bench & 80.30 & 56.20 & 80.10 & \textbf{85.40} & 78.79 \\
 & ACEBench-en & 81.40 & 80.90 & 79.30 & 80.90 & \textbf{87.70} \\
 & ACEBench-zh & 83.40 & 87.50 & 83.60 & 85.00 & \textbf{89.60} \\
 & BFCL v4 & 61.18 & 52.30 & 61.60 & \textbf{68.14} & 66.47 \\
 & BrowseComp-zh & \textbf{65.00} & 28.70 & 61.90 & 63.67 & 64.71 \\
 & SpreadSheetBench & 35.29 & 27.70 & 34.00 & \textbf{55.36} & 40.08 \\
\bottomrule
\end{tabular}%
}
\caption{Evaluation of post-trained models across a wide range of language benchmarks.}
\label{tab:posttrain-eval-text}
\end{table*}

\paragraph{Evaluation of Post-trained Models.}

As shown in Table~\ref{tab:posttrain-eval-text}, the post-trained \ourmodel achieves competitive or leading performance across a broad set of text-centric benchmarks, matching strong open-source and proprietary models on knowledge, instruction-following, coding, and agent-oriented tasks, while maintaining strong general reasoning ability, despite being a natively unified omni model:

On knowledge-intensive benchmarks, such as SimpleQA and ChineseSimpleQA, \ourmodel further improves upon the already strong pre-training results. This indicates that post-training effectively enhances factual recall and answer calibration, while preserving the underlying knowledge representations learned during large-scale pre-training.

On general reasoning, mathematical, and coding benchmarks, \ourmodel demonstrates stable and competitive performance against strong post-trained baselines. Although Gemini~3-Pro achieves leading results on several particularly challenging benchmarks, including GPQA-Diamond, AIME~2025, HMMT~2025, and LiveCodeBench, \ourmodel consistently matches or outperforms DeepSeek-V3.2-Thinking (DS-V3.2-Thinking), Gemini~2.5-Pro, and GPT-5 (High) on most of the benchmarks. This behavior aligns well with the pre-training observations, suggesting that \ourmodel emphasizes robust and balanced capability development rather than aggressive optimization toward extreme reasoning or competition-style tasks.

On instruction-following benchmarks, \ourmodel shows clear advantages on multi-instruction evaluations, achieving the best performance on MultiChallenge and Multi-IF, as well as near-top results on IFEval. These results suggest that post-training effectively strengthens instruction compliance and compositional instruction understanding, complementing the strong pre-trained foundation.

On agent-oriented benchmarks, \ourmodel demonstrates competitive and often leading performance, particularly on ACEBench (in both English and Chinese) and BrowseComp-zh. While Gemini~3-Pro excels on certain tool-intensive tasks, \ourmodel exhibits strong generalization across diverse agent scenarios, indicating practical usability in complex interactive environments.

The comparison between pre-training and post-training results highlights a consistent capability trajectory of \ourmodel. Strengths in factual knowledge and general robustness established during pre-training are almost retained after post-training, while instruction following and agent capabilities are significantly enhanced. Although \ourmodel remains competitive across general reasoning, math, and coding tasks, a moderate gap persists on the most challenging reasoning benchmarks compared to models such as Gemini~3-Pro. Our future wok will further leverage architectural design and advanced training strategies to better support complex, long-horizon reasoning.

\subsection{Evaluation on Vision Benchmarks}
\label{sec:eval-visual}

To evaluate the visual understanding and generation capabilities enabled by native multimodal training, we assess \ourmodel on various image-centric and video-centric benchmarks. These evaluations cover visual reasoning, document understanding, visual question answering, video understanding, as well as image and video generation, providing a holistic assessment of how the unified multi-modal architecture manifests as performances across perception, reasoning, and generation tasks.

\paragraph{Visual Understanding \& Generation Benchmarks}
\begin{itemize}
    \item \textbf{STEM and Reasoning: } MMMU-Pro~\citep{mmmupro}, MathVista~\citep{mathvista}, MathVerse~\citep{zhang2024mathverse}, MathVision~\citep{mathvision}, VisualPuzzle~\citep{visualpuzzles}, VisuLogic~\citep{xu2025visulogic}, VLMAreBlind~\citep{vlmareblind}, MMMU~\citep{mmmu}.
    \item \textbf{Document Understanding: }ChartQA~\citep{chartqa}, AI2D~\citep{ai2d}, DocVQA(val) \citep{docvqa}, OCRBench~\citep{ocrbench}, ChartXiv-RQ, ChartXiv-DQ~\citep{wang2024charxiv}.
    \item \textbf{General VQA: } SimpleVQA~\citep{simplevqa}, HallusionBench~\citep{guan2024hallusionbench}, MMStar~\citep{mmstar}, BLINK~\citep{fu2024blink}, CV-Bench~\citep{zhu2025cvbench}, CountBench~\citep{countbench}.
    \item \textbf{Video Understanding: } VideoMME~\citep{videomme}, Video-MMMU~\citep{videommmu}, MMVU~\citep{zhao2025mmvu}.
     \item \textbf{Image and Video Generation: }GenEval~\citep{ghosh2023geneval}, VBench~\citep{huang2024vbench}.
\end{itemize}

\paragraph{Evaluation of Visual Understanding}

\begin{table}[t]
  \centering
\footnotesize
\setlength{\tabcolsep}{2.1pt}
    \resizebox{\textwidth}{!}{
    \begin{tabular}{llccccc}
    \toprule
    \textbf{Category} & \textbf{Benchmark} & \textbf{Qwen3-VL Thinking} & \textbf{Gemini 2.5-Pro} & \textbf{GPT-5 (High)} & \textbf{Gemini 3-Pro} & \textbf{\ourmodel} \\
    \midrule
    \multirow{7}{*}{\shortstack[l]{STEM\\\&\\Reasoning}}
      & MMMU-Pro & 68.28 & 68.80 & 78.40 & \textbf{81.00} & 68.63 \\
      & MathVista & 86.80 & 82.70 & 82.10 & \textbf{89.20} & 84.80 \\
      & MathVerse & 83.96 & 86.27 & 84.19 & \textbf{91.62} & 85.13 \\
      & MathVision & 71.84 & 73.30 & 78.06 & \textbf{87.27} & 74.34 \\
      & VisualPuzzle & 57.01 & 61.51 & 57.75 & \textbf{71.48} & 64.82 \\
      & VisuLogic & 31.93 & 32.80 & 29.80 & \textbf{37.60} & 32.00 \\
      & VLMAreBlind & 75.13 & 76.54 & 69.60 & 80.83 & \textbf{91.38} \\
    \midrule
    \multirow{6}{*}{\shortstack[l]{Document\\Understanding}} 
      & ChartQA & 84.60 & 84.08 & 78.24 & \textbf{89.44} & 87.80 \\
      & AI2D & 96.96 & 97.09 & 95.63 & \textbf{97.70} & 96.89 \\
      & DocVQA$_{\textit{val}}$ & 95.44 & 91.43 & 94.16 & 90.70 & \textbf{95.45} \\
      & OCRBench & 863& 866& 804& \textbf{909}& 878\\
      & ChartXiv-RQ & 63.00 & 67.80 & 81.10 & \textbf{81.40} & 67.10 \\
      & ChartXiv-DQ & 92.92 & 93.38 & 91.17 & \textbf{95.95} & 89.05 \\
    \midrule
    \multirow{6}{*}{General VQA} 
      & SimpleVQA & 62.37 & 68.19 & 55.84 & \textbf{74.06} & 67.64 \\
      & HallusionBench & 64.01 & 63.70 & 66.58 & \textbf{73.48} & 63.87 \\
      & MMStar & 76.88 & 77.50 & 82.10 & \textbf{82.96} & 75.54 \\
      & BLINK & 66.60 & 70.60 & 70.39 & \textbf{77.49} & 70.02 \\
      & CV-Bench & 87.57 & 84.87 & 84.99 & \textbf{90.07} & 87.19 \\
      & CountBench & 92.67 & 91.00 & 88.88 & \textbf{97.35} & 96.54 \\
    \midrule
    \multirow{3}{*}{\shortstack[l]{Video\\Understanding}} 
      & VideoMME$_{\textit{(w sub)}}$ & 80.97 & 86.90 & 87.36 & \textbf{88.40} & 81.35 \\
      & Video-MMMU & 80.00 & 83.60 & 84.60 & \textbf{87.60} & 81.11 \\
      & MMVU & 71.10 & 76.10 & \textbf{87.34} & 76.30 & 72.24 \\
    \bottomrule
    \end{tabular}%
    }
    \caption{Evaluation of post-trained models across a wide range of vision benchmarks.}
    \label{tab:posttrain-eval-multimodal}
\end{table}

\begin{table}[t]
  \centering
\footnotesize
    \resizebox{\textwidth}{!}{%
    \begin{tabular}{lccccccc}
        \toprule
        \textbf{Model} & \textbf{MathVista} & \textbf{MathVerse} & \textbf{MathVision} & \textbf{VisualPuzzle} & \textbf{VisuLogic}  & \textbf{ChartQA} & \textbf{AI2D} \\
        \midrule
        \multirow{4}{*}{\textbf{\basemodel}}
        & 84.40 & 81.45 & 68.75 & 54.24 & 28.50 & 87.68 & 96.02 \\
        \cmidrule(l){2-8} 
         & \textbf{OCRBench} & \textbf{ChartXiv-RQ} & \textbf{ChartXiv-DQ} & \textbf{SimpleVQA} & \textbf{MMStar} & \textbf{BLINK} & \textbf{CV-Bench} \\
        \cmidrule(l){2-8} 
         & 875 & 62.70 & 90.35 & 61.54 & 74.07 & 64.12 & 86.96 \\
        \bottomrule
    \end{tabular}%
    }
    \caption{Evaluation of our pre-trained model across a wide range of vision benchmarks, There are few publicly available visual base models and their results, here we only report the results of our model.}
    \label{tab:posttrain-eval-multimodal-base}
\end{table}

\begin{table}[t]
    \centering
    \footnotesize
    \begin{tabular}{lcccccc}
        \toprule
        \multirow{1}{*}{\textbf{Benchmark}} & \textbf{Nano Banana Pro} & \textbf{Seedream 4.0} & \textbf{GPT-Image} & \textbf{Qwen-Image} & \textbf{\basemodel} & \textbf{\ourmodel} \\
        \midrule
        GenEval & 89.0 & 85.4 & 84.0 & \textbf{91.0} & 88.4 & 90.1 \\
        \bottomrule
    \end{tabular}
    \caption{Comparison of the image generation ability on GenEval against specialized models.}
    \label{tab:pretrain-eval-imagegen}
\end{table}

\begin{table}[H]
    \centering
    \footnotesize
    \begin{tabular}{lcccccc}
        \toprule
        \textbf{Benchmark} & \textbf{Metric} 
        & \textbf{HunyuanVideo-1} 
        & \textbf{Wan2.1-14B-0725} 
        & \textbf{Veo3} 
        & \textbf{\basemodel} 
        & \textbf{\ourmodel} \\
        \midrule
        \multirow{3}{*}{VBench} 
        & Quality   & 85.07 & 85.59 & \textbf{85.70} & 84.14 & 84.40 \\
        & Semantic  & 76.88 & 76.11 & 82.49 & 82.31 & \textbf{83.40} \\
        & Overall   & 83.43 & 83.69 & \textbf{85.06} & 83.78 & 84.20 \\
        \bottomrule
    \end{tabular}
    \caption{Comparison of the video generation ability on VBench against specialized models.}
    \label{tab:pretrain-eval-videogen}
\end{table}

Table~\ref{tab:posttrain-eval-multimodal} and Table~\ref{tab:posttrain-eval-multimodal-base} illustrate the multimodal capabilities of \ourmodel under the instruction-tuned setting (\ourmodel) and its native multimodal foundation (\basemodel), respectively. This comparison enables a clearer understanding of how instruction tuning interacts with a unified multimodal pre-training paradigm.

\basemodel exhibits strong and well-rounded performances across visual reasoning, document understanding, and general VQA benchmarks, despite the absence of instruction tuning. 
It suggests that core multimodal perception and cross-modal reasoning abilities are largely acquired during pre-training, benefiting from early fusion of visual and textual tokens and modality-agnostic representation learning. This behavior differs from modular or late-fusion designs, where comparable capabilities typically emerge only after extensive task-specific tuning.
Based on this foundation, \ourmodel consistently improves performance on most benchmarks, particularly on tasks requiring explicit reasoning, compositional understanding, and robust visual–language alignment.

Compared with other strong multimodal baselines, \ourmodel achieves competitive or superior results on a wide range of reasoning, document understanding, and general VQA tasks, while maintaining balanced performance rather than optimizing for individual benchmarks. It indicates that instruction tuning in \ourmodel primarily refines reasoning and alignment behaviors, instead of compensating for perceptual capacity.

In document understanding scenarios, the relatively strong results of \basemodel already demonstrate effective layout-aware reasoning and text extraction, while \ourmodel further enhances question understanding and structured reasoning over complex visual documents. A similar trend is observed in general VQA and video understanding benchmarks, where instruction tuning improves robustness and temporal reasoning without altering the underlying architecture.

\paragraph{Evaluation of Visual Generation}

We evaluate \ourmodel on widely used benchmarks for both image and video generation, and compare it with leading commercial and open-source models.

For image generation, as shown in Table~\ref{tab:pretrain-eval-imagegen}, \ourmodel achieves competitive performance on the GenEval benchmark. In particular, \ourmodel performs on par with state-of-the-art commercial systems such as Nano-Banana Pro~\citep{gemini3.0} and Qwen-Image~\citep{qwen-image}, and is comparable to GPT-Image~\citep{gpt4o} and Seedream~4.0~\citep{seedream4}. 
It demonstrates that \ourmodel is capable of producing high-aesthetic images with strong semantic alignment and fine-grained visual details.

For video generation, Table~\ref{tab:pretrain-eval-videogen} summarizes the results on the VBench benchmark. \ourmodel achieves the best performance on VBench-Semantic, surpassing strong commercial models such as Veo3~\citep{veo3}, which indicates its superior semantic alignment in video generation. This advantage aligns with the unified multimodal architecture, where high-level semantic representations are effectively transferred to generative tasks. Meanwhile, \ourmodel remains competitive on the overall and quality metrics, demonstrating solid visual fidelity and temporal consistency. 
In addition, \ourmodel performs on par with leading open-source models such as HunyuanVideo-1~\citep{hunyuanvideo} and Wan2.1-14B-0725~\citep{wan}, indicating that the proposed architecture provides a strong video generation foundation even before post-training. Overall, these results validate the effectiveness of \ourmodel in producing semantically accurate and visually coherent videos.

\subsection{Evaluation on Audio Benchmarks}
\label{sec:eval-audio}

To evaluate the audio understanding and generation capabilities enabled by native multimodal training, we assess \ourmodel on a diverse suite of speech- and audio-centric benchmarks. These evaluations cover automatic speech recognition, speech-based dialogue, general audio understanding, as well as speech generation, providing a comprehensive view of how the unified multimodal architecture translates into robust audio perception, semantic understanding, and generation performance across across diverse linguistic and acoustic environments.

\paragraph{Audio Understanding \& Generation Benchmarks}
\begin{itemize}
    \item \textbf{Automatic Speech Recognition}: AISHELL-1~\citep{bu2017aishell}, AISHELL-2~\citep{du2018aishell}, WenetSpeech~\citep{zhang2022wenetspeech}, LibriSpeech~\citep{panayotov2015librispeech}, Fleurs~\citep{conneau2023fleurs}.
    \item \textbf{Voice Chatting}: VoiceBench~\citep{chen2024voicebench}.
    \item \textbf{Audio Understanding}: MMAU~\citep{sakshi2024mmau}, CochlScene~\citep{jeong2022cochlscene}, TUT2017 \citep{mesaros2016tut},  ClothoAQA~\citep{lipping2022clotho}, VocalSound~\citep{gong2022vocalsound}.
    \item \textbf{Speech Generation}: Seed-TTS~\citep{seed-tts}.
\end{itemize}

\begin{table*}[t]
\centering

\small
\setlength{\tabcolsep}{3.0pt}
\renewcommand{\arraystretch}{1.08}

\begin{adjustbox}{max width=\textwidth}
\begin{tabular}{lcccccccc}
\toprule
\textbf{Benchmark}
 & \makecell{\textbf{Kimi}\\\textbf{Audio}}
 & \makecell{\textbf{GPT-4o}\\\textbf{-Audio}}
 & \makecell{\textbf{Qwen3-Omni}\\\textbf{-Instruct}}
 & \makecell{\textbf{LongCat-Flash}\\\textbf{-Omni}}
 & \makecell{\textbf{Gemini-3}\\\textbf{-Pro}}
 & \makecell{\textbf{\basemodel}} 
 & \makecell{\textbf{\ourmodel}} \\
\midrule

\multicolumn{8}{c}{\textit{Automatic Speech Recognition ($\downarrow$})} \\
\midrule
AISHELL-1
& 0.60 & 3.52 & 0.84 & 0.63 & 3.04 & 0.75 & \textbf{0.31} \\

AISHELL-2
& 2.56 & 4.26 & 2.34 & 2.78 & 4.98 & 2.90 & \textbf{2.64} \\

\makecell[l]{WenetSpeech\\ \textit{net $|$ meeting}}
& 5.37 $|$ 6.28
& 15.30 $|$ 32.27
& \textbf{4.69} $|$ \textbf{5.89}
& 6.09 $|$ 6.69
& 10.94 $|$ 12.08
& 11.57 $|$ 22.85
& 7.27 $|$ 7.36 \\

\makecell[l]{LibriSpeech\\ \textit{clean $|$ other}}
& 1.28 $|$ \textbf{2.42}
& 1.39 $|$ 3.75
& 1.22 $|$ 2.48
& 1.57 $|$ 4.01
& 2.74 $|$ 4.40
& 1.47 $|$ 3.73
& \textbf{1.16} $|$ 2.61 \\

Fleurs-en
& 4.44 & 3.32 & \textbf{2.72} & 5.02 & 3.63 & 4.39 & 3.14 \\

Fleurs-zh
& 2.69 & 2.44 & 2.20 & 3.99 & 4.98 & 1.58 & \textbf{0.83} \\

\midrule
\multicolumn{8}{c}{\textit{VoiceBench} ($\uparrow$) } \\
\midrule
AlpacaEval
& 4.46 & 4.73 & 4.74 & 4.94 & \textbf{4.80} & 4.65 & 4.62 \\

CommonEval
& 3.97 & 4.37 & 4.54 & 4.32 & \textbf{4.68} & 4.39 & 3.74 \\

SD-QA
& 63.12 & 90.10 & 76.90 & 82.46 & \textbf{94.39} & 86.44 & 77.58 \\

MMSU
& 62.17 & 78.90 & 69.00 & 81.95 & \textbf{92.16} & 76.61 & 84.68 \\

OpenBookQA
& 83.52 & 87.90 & 89.70 & 93.41 & \textbf{96.26} & 88.35 & 92.97 \\

IFEval
& 61.10 & 66.81 & 77.80 & 77.99 & \textbf{87.45} & 71.17 & 72.67 \\

AdvBench
& \textbf{100.00} & 99.23 & 99.30 & \textbf{100.00} & 98.46 & 98.65 & 99.23 \\

\midrule
\multicolumn{8}{c}{\textit{Audio Understanding} ($\uparrow$) } \\
\midrule
MMAU
& 65.20 & 68.40 & 77.50 & 75.90 & \textbf{80.80} & \textbf{80.80} & 80.40 \\

TUT2017
& 65.25 & 20.74 & 40.74 & 65.43 & 61.42 & 57.65 & \textbf{68.09} \\

CochlScene
& 80.42 & 34.94 & 43.03 & 70.02 & 74.60 & 75.24 & \textbf{82.77} \\

ClothoAQA
& 72.21 & 61.87 & \textbf{75.16} & 72.83 & 74.41 & 65.70 & 73.68 \\

VocalSound
& \textbf{94.85} & 82.37 & 91.60 & 92.76 & 92.01 & 91.48 & 90.73 \\

\bottomrule
\end{tabular}
\end{adjustbox}

\caption{Comparison of the automatic speech recognition, speech-to-text dialogue, and audio understanding abilities against specialized models.
Automatic speech recognition uses word error rate (WER) as the metric (lower is better), while other tasks use accuracy or score (higher is better).}
\label{tab:audio_und}
\end{table*}

\paragraph{Evaluation of Audio Understanding}

We compare \ourmodel with other leading specialist and generalist models on ASR, voice-chatting, and other audio understanding benchmarks.
As shown in Table~\ref{tab:audio_und} and Table~\ref{tab:audio_tts}, \ourmodel exihibits state-of-the-art or competitive performance across speech recognition, speech-to-text dialogue, audio understanding and speech generation benchmarks.

On automatic speech recognition (ASR) tasks, \ourmodel achieves low word error rate (WER) on both Chinese and English benchmarks, including AISHELL and LibriSpeech, indicating robust performance across languages and acoustic conditions. While some models achieve stronger results on specific datasets such as WenetSpeech, \ourmodel maintains consistently stable performance across a wide range of ASR benchmarks, reflecting good generalization.

On VoiceBench, \ourmodel remains competitive on tasks such as MMSU and OpenBookQA, reflecting its ability to handle speech-based interaction and knowledge-grounded reasoning.

For audio understanding tasks, \ourmodel performs favorably on acoustic scene and environmental sound benchmarks such as TUT2017 and CochlScene, while achieving comparable results on more diverse benchmarks including MMAU and ClothoAQA. Overall, the model shows consistent performance across a wide range of non-speech and general audio understanding tasks, indicating its ability to capture audio semantics beyond speech content.

\paragraph{Evaluation of Audio Generation}

\begin{table}[t]
\centering
\small
\setlength{\tabcolsep}{8pt}
\begin{tabular}{l l c}
\toprule
\textbf{Benchmark} & \textbf{Model} & \textbf{Performance} \\
\midrule
\multirow{12}{*}{\shortstack[l]{SEED-TTS ($\downarrow$)\\  \textit{test-zh $|$ test-en}}}
& Seed-TTS$_{\text{ICL}}$  & 1.11 $|$ 2.24 \\
\makecell[l]{}
& Seed-TTS$_{\text{RL}}$  & 1.00 $|$ 1.94 \\
\makecell[l]{}
& MaskGCT  & 2.27 $|$ 2.62 \\
\makecell[l]{}
& E2 TTS  & 1.97 $|$ 2.19 \\
\makecell[l]{}
& F5-TTS  & 1.56 $|$ 1.83 \\
\makecell[l]{}
& Spark TTS  & 1.20 $|$ 1.98 \\
\makecell[l]{}
& CosyVoice 2  & 1.45 $|$ 2.57 \\
\makecell[l]{}
& CosyVoice 3  & \textbf{0.71} $|$ 1.45 \\
\makecell[l]{}
& Qwen2.5-Omni  & 1.42 $|$ 2.33 \\
\makecell[l]{}
& Qwen3-Omni & 1.07 $|$ \textbf{1.39} \\
\makecell[l]{}
& \textbf{\basemodel}  & 3.41 $|$ 2.44 \\
\makecell[l]{}
& \textbf{\ourmodel}  & 1.35 $|$ 1.54 \\
\bottomrule
\end{tabular}
\caption{
Comparison of the text-to-speech ability on SEED-TTS against specialized models. Word Error Ratio (WER) is used to evaluate content consistency, the lower the better.
}
\label{tab:audio_tts}
\end{table}

On SEED-TTS, \ourmodel achieves competitive content consistency on both Chinese and English test sets (test-zh, test-en), performing comparably to recent audio–language models such as Qwen3-Omni. While specialist TTS systems (e.g., CosyVoice-3) obtain stronger results, \ourmodel demonstrates reliable content preservation without task-specific TTS optimization.

Taken together, the results in Table~\ref{tab:audio_und} indicate that ERNIE-5.0 is a competitive unified audio–language model that balances speech recognition accuracy, speech interaction, general audio understanding and speech generation within a single framework.

\subsection{Discussion}
\label{sec:pretrain-discuss}

To better understand the behavior of \ourmodel in large-scale multimodal training, we dissect two pivotal architectural designs: modality-agnostic expert routing and elastic training. The following subsections analyze these two components and their effects on efficiency and scalability.

\subsubsection{Modality-Agnostic Expert Routing}
\label{sec:moe-routing-discussion}
\ourmodel employs a modality-agnostic expert routing mechanism, without introducing modality-specific parameters or routing rules. All inputs, including text, image, video, and audio, are processed by the same routing network and share a common pool of experts.
In this discussion, we examine the expert routing behavior of the MoE model, with particular attention to modality-specific utilization patterns, cross-modality expert overlap, and routing balance across layers.

\begin{figure}[t]
  \centering
  \begin{subfigure}[b]{0.32\textwidth} 
    \centering
    \includegraphics[width=\linewidth]{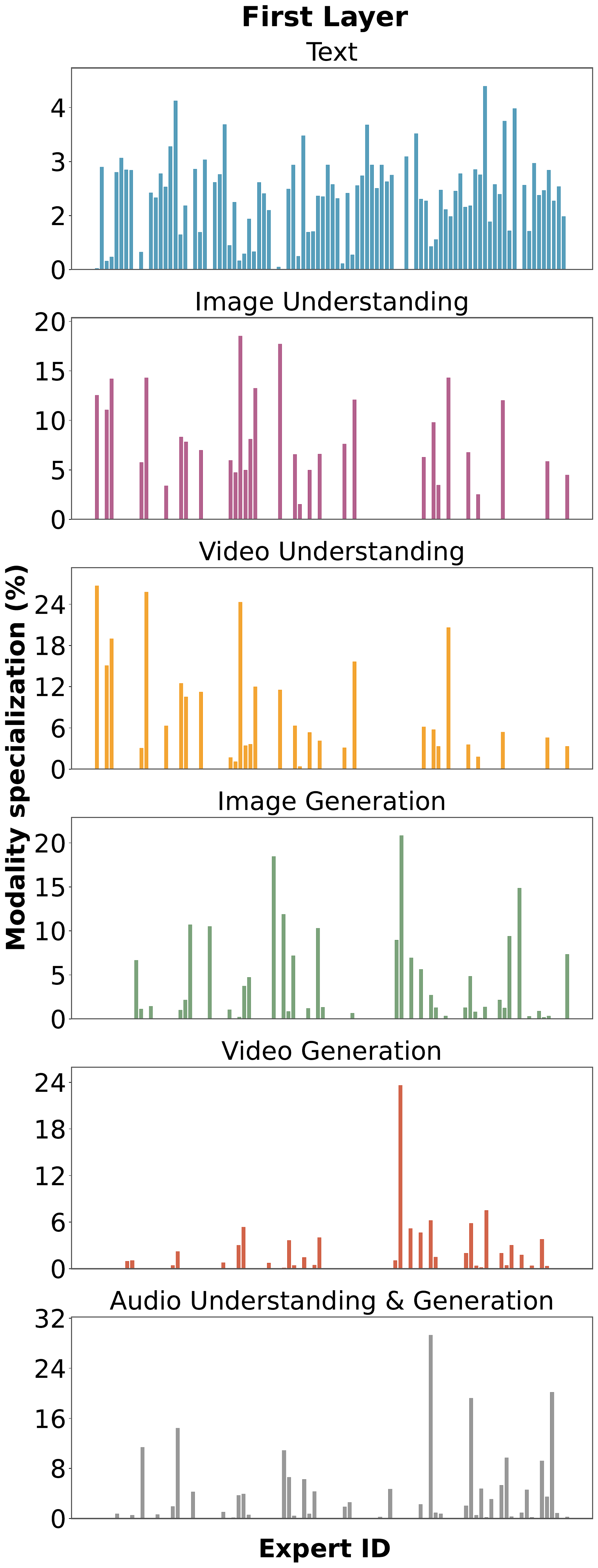} 
  \end{subfigure}
  \hfill
  \begin{subfigure}[b]{0.32\textwidth}
    \centering
    \includegraphics[width=\linewidth]{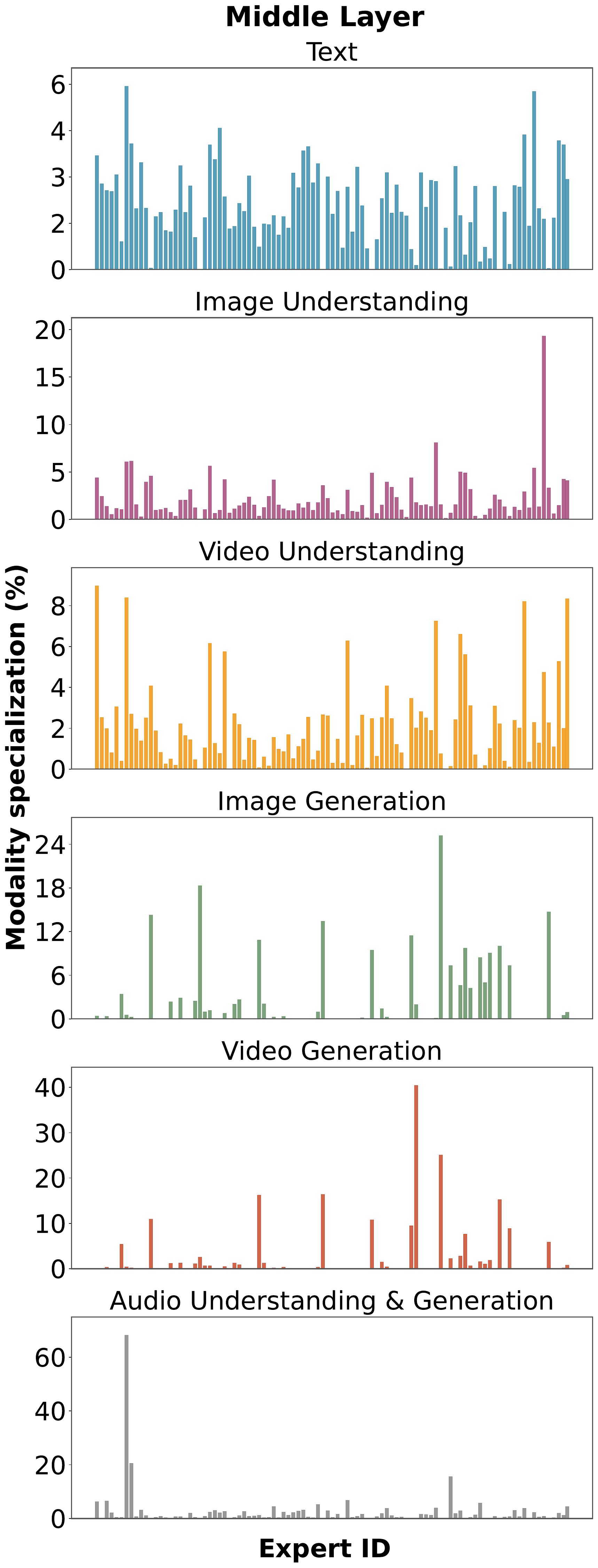}
  \end{subfigure}
  \hfill
  \begin{subfigure}[b]{0.32\textwidth}
    \centering
    \includegraphics[width=\linewidth]{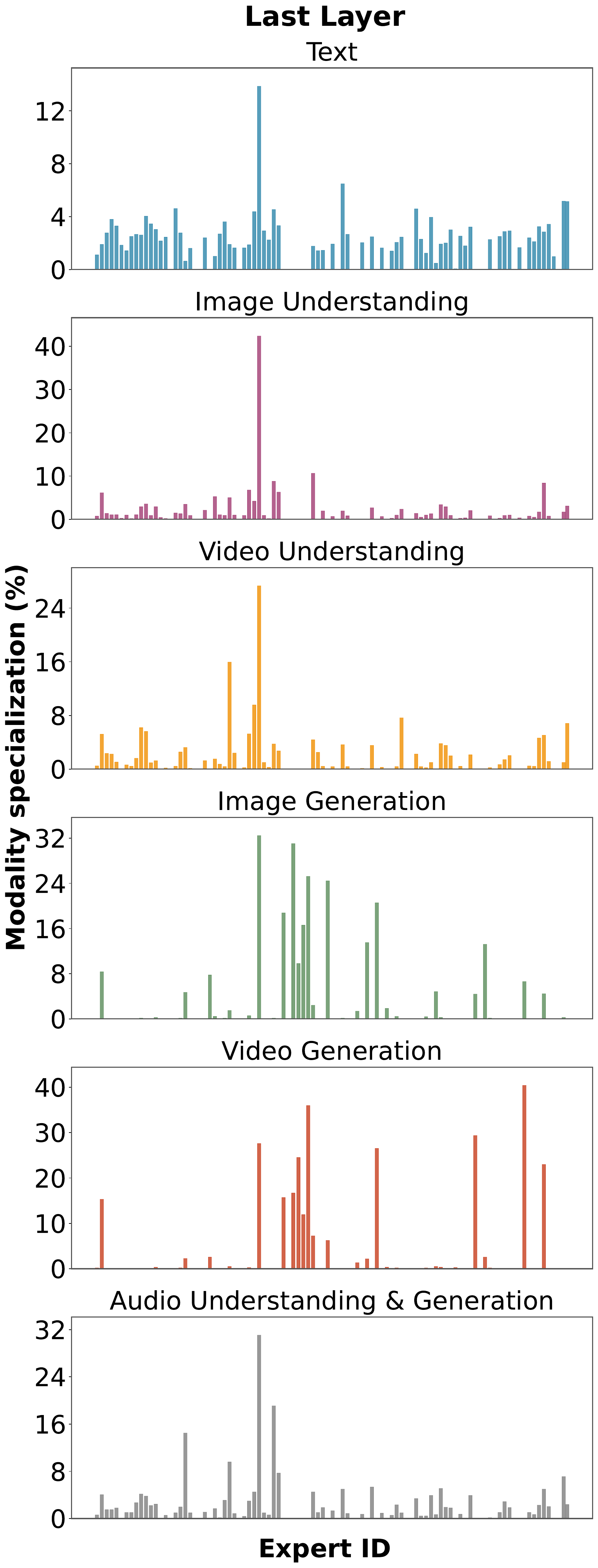}
  \end{subfigure}
  \caption{
  Visualization of expert utilization patterns across modalities and tasks for the representative first, middle and last layers. The y-axis indicates the frequency of expert activation.}
  \label{fig:moe-routing-vis}
\end{figure}

\paragraph{Perspective of Expert Utilization and Specialization} 
Figure~\ref{fig:moe-routing-vis} reveals a clearly non-uniform expert activation distribution, suggesting that different experts still play distinct functional roles under the modality-agnostic expert routing setting.
A subset of experts is repeatedly activated across text, image, video, and audio inputs, whereas the remaining experts exhibit strong modality-specific activation patterns.
Expert activations for image, video, and audio inputs are noticeably more concentrated than those for text-only inputs.
From a task-level perspective, visual generation and audio-related tasks lead to more concentrated expert activations than text and visual understanding tasks.

\begin{figure}[t]
  \centering
  \begin{subfigure}[b]{0.325\textwidth} 
    \centering
    \includegraphics[width=\linewidth]{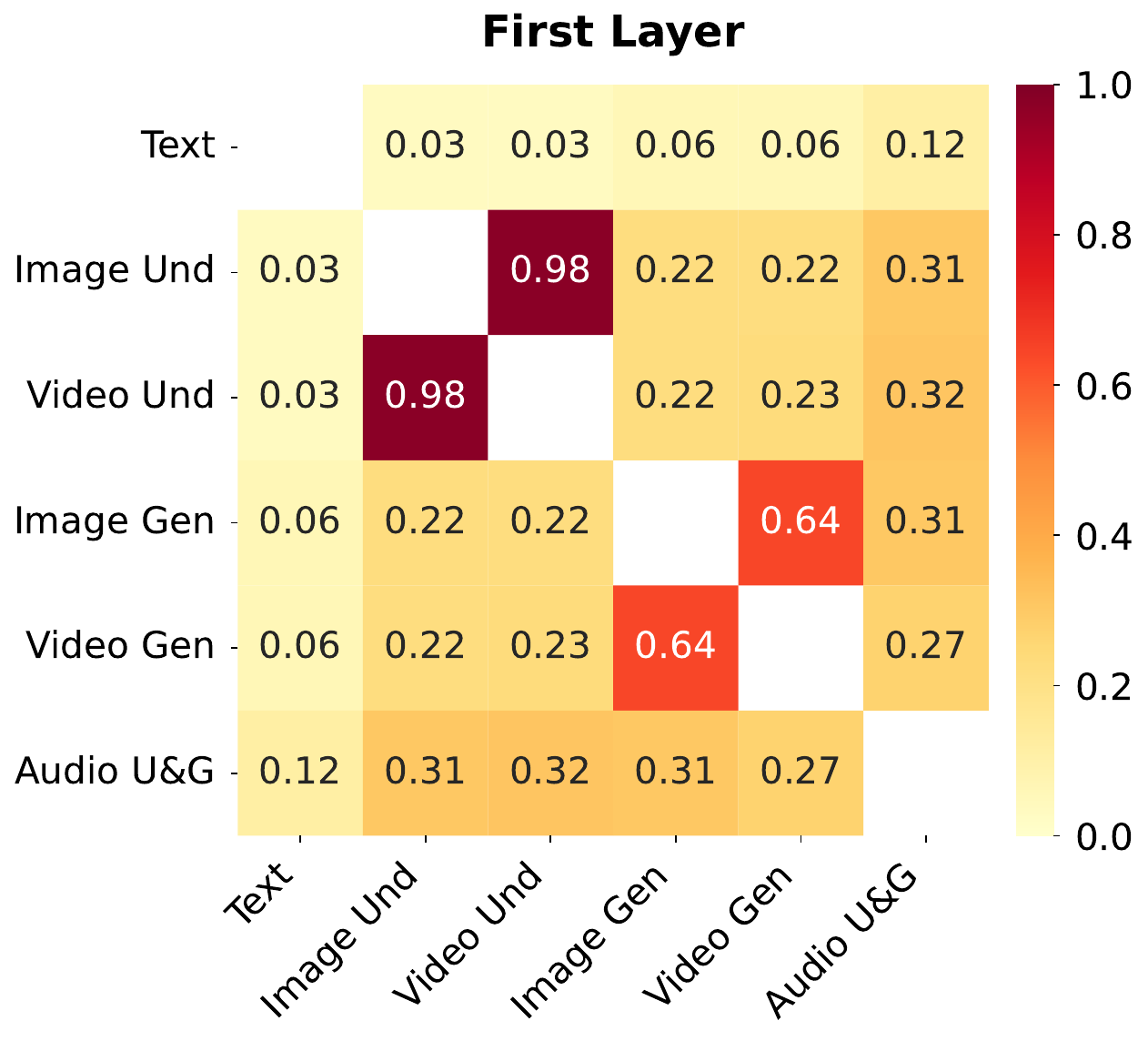} 
  \end{subfigure}
  \hfill
  \begin{subfigure}[b]{0.325\textwidth}
    \centering
    \includegraphics[width=\linewidth]{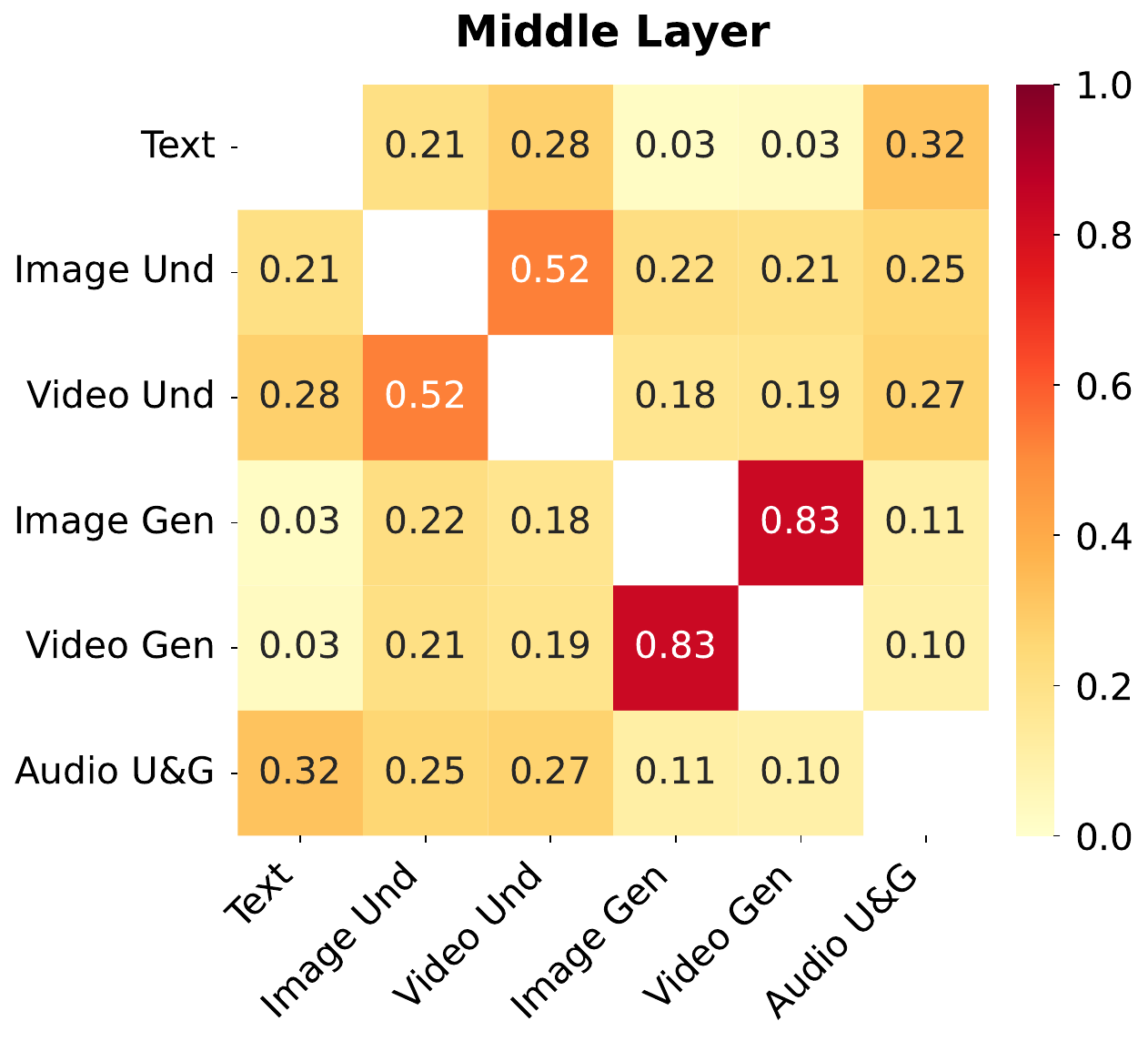}
  \end{subfigure}
  \hfill
  \begin{subfigure}[b]{0.325\textwidth}
    \centering
    \includegraphics[width=\linewidth]{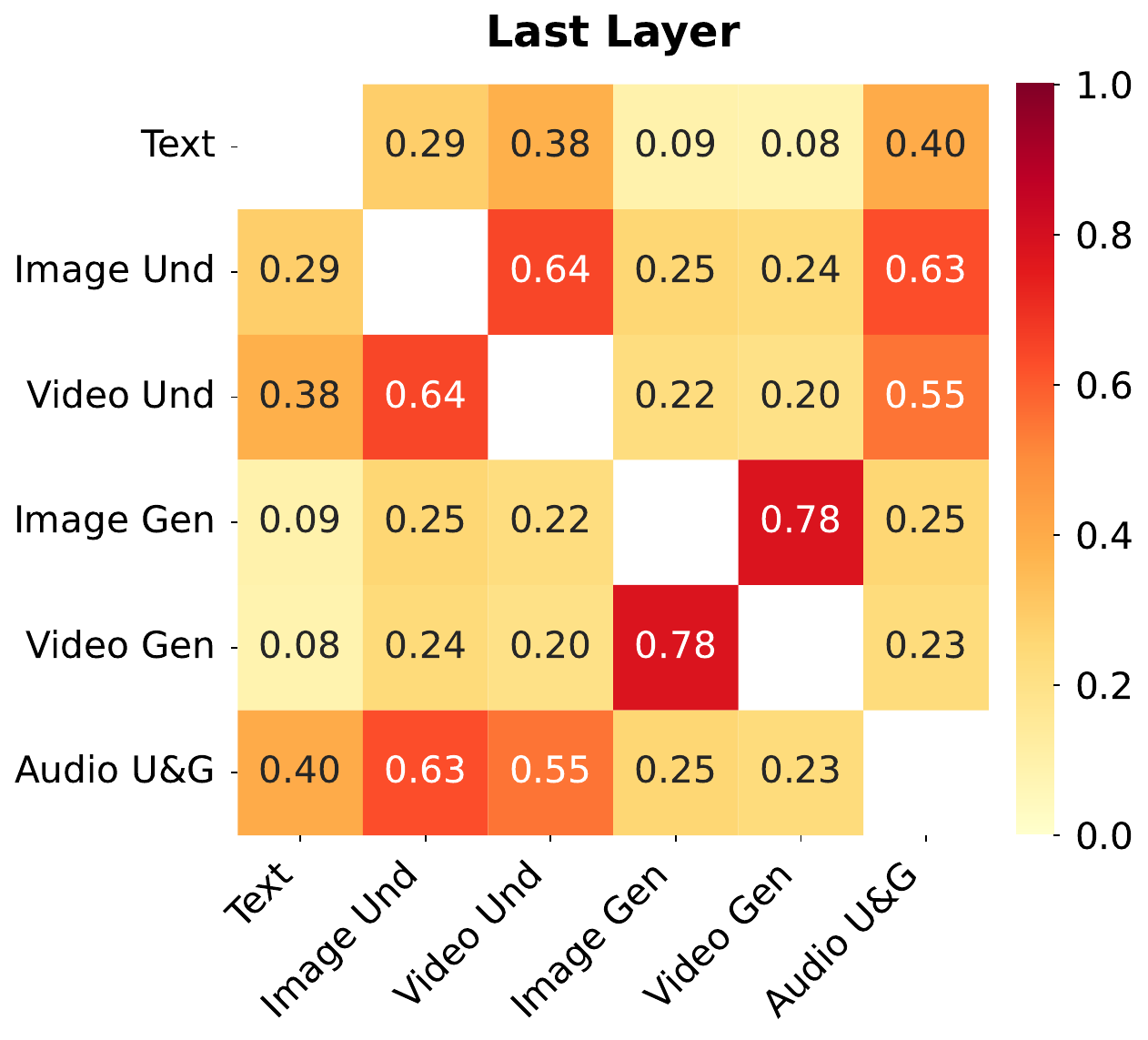}
  \end{subfigure}
  \caption{Visualization of expert collaboration across modalities and tasks for the representative first, middle and last layers, using the Intersection over Union (IoU) of the top 25\% most frequently activated experts for each modality.}
  \label{fig:hot-experts-IoU}
\end{figure}

\paragraph{Perspective of Cross-Modality Expert Collaboration} 
Figure~\ref{fig:hot-experts-IoU} further analyzes cross-modality expert collaboration by reporting the Intersection over Union (IoU) of the top 25\% of experts by activation frequency for each modality across different layers. 
From the overlap patterns, text shows a higher degree of co-activation with audio than with other modalities, and this overlap with image and video becomes increasingly pronounced in deeper layers.
This trend suggests that multimodal representations gradually shift from low-level modality-specific features toward higher-level unified semantic features, enabling stronger collaboration with text at deeper layers.
For visual modalities, image and video understanding tasks exhibit high expert overlap, and a similar pattern is observed for image and video generation tasks, which aligns with the architectural design in which an image is regarded as a single-frame video.
In contrast, the overlap between visual understanding and generation remains relatively low, with no clear preference toward either direction.

\begin{figure}[t]
  \centering
    \includegraphics[width=0.58\linewidth]{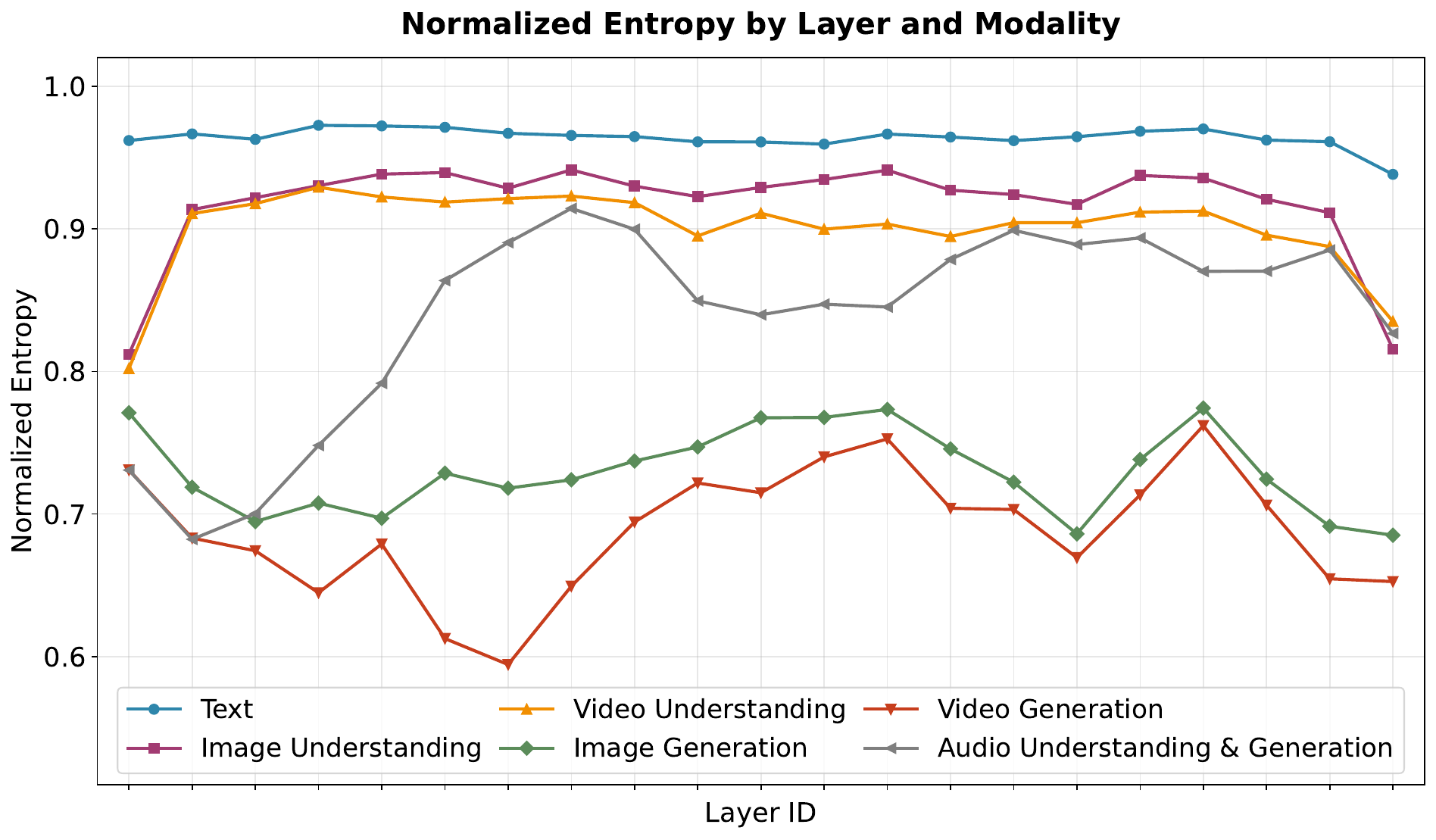} 
    \caption{Visualization of the load balancing across layers and modalities, using the normalized entropy (NE) metric of expert routing.}
    \label{fig:moe_experts_normalized_entropy}
\end{figure}

\paragraph{Perspective of Load Balancing}
Figure~\ref{fig:moe_experts_normalized_entropy} reports the normalized entropy (NE) of expert routing across layers and modalities, defined as $\mathrm{NE} = \frac{-\sum_{i=1}^{N} p_i \log(p_i)}{\log N}$, which provides a quantitative measure of expert load balance.
Here, $N$ is the number of experts and $p_i$ is the fraction of tokens routed to $i$-th expert. The value of $\mathrm{NE}$ lies in $[0,1]$, where larger values correspond to more uniform expert utilization.

From a layer-wise perspective, the text modality exhibits consistently high and stable normalized entropy across almost all layers, with only a mild drop observed at the final layer. Notably, the first layer does not show severe imbalance, contradicting the common assumption that early MoE layers require dense designs to maintain routing stability~\citep{deepseekv3}.

For visual understanding, expert utilization becomes less balanced at both the lowest and highest layers, while the middle layers maintain relatively uniform routing, suggesting that visual perception benefits from balanced expert sharing during intermediate feature abstraction stages.
In contrast, visual generation and audio-related tasks display a different trend: the first layer shows a moderate level of balance, followed by a decrease in entropy in lower layers, a partial recovery in lower-to-mid layers, and a fluctuating drop at the higher layers. This pattern indicates alternating phases of expert specialization and re-integration along depth for generative and audio-centric tasks.

Taken together, these analyses reveal clear expert utilization patterns across layers, modalities, and tasks in \ourmodel.
Although the routing mechanism is modality-agnostic during training, pronounced expert specialization across modalities still emerges, which suggests that the router could capture modality structure and allocate expert capacity in a self-driven manner.
From a system perspective, the unified routing mechanism simplifies the overall design and avoids manual expert partitioning across modalities.
Beyond empirical observations, the routing behaviors and load-balancing characteristics also offer practical guidance for future model design.
For example, layer-aware expert allocation, adaptive balancing strategies, and modality-shared expert configurations may have the potential to become effective principles for building the next generation of native multimodal architectures.

\begin{table}[t]
\centering
\small
\begin{tabular}{lcc}
\toprule
\textbf{Training Configuration} & \textbf{Inference Configuration (\# of Layers)} & \textbf{Validation Loss} \\
\midrule
Baseline{\tiny Layers=16} & Layers=16 & 1.945 \\
\midrule
\multirow{2}*{Elastic Depth{\tiny Layers $\in [1, 16]$}} 
& Layers=16 & 1.941 \\
& Layers=12 & 2.137 \\
\bottomrule
\end{tabular}
\caption{Validation loss under elastic depth training and inference configurations for small-scale MoE model. Elastic depth slightly improves full-depth performance and yields reduced-depth sub-networks with predictable degradation, enabling flexible deployment without retraining.}
\label{tab:elastic-depth}
\end{table}

\begin{table}[t]
\centering
\small
\begin{tabular}{lcc}
\toprule
\textbf{Training Configuration} & \textbf{Inference Configuration (\# of Experts)} & \textbf{Validation Loss} \\
\midrule
Baseline{\tiny Experts=64} & Experts=64 & 1.957 \\
\midrule
\multirow{2}*{Elastic Width{\tiny Experts $\in \{64, 32\}$}} 
& Experts=64 & 1.964 \\
& Experts=32 & 2.218 \\
\bottomrule
\end{tabular}
\caption{Validation loss under elastic width configurations for the
small-scale MoE model. Elastic width introduces only minor degradation at full capacity, while reduced-width models remain usable, supporting deployment under constrained parameter budgets.}
\label{tab:elastic-width}
\end{table}

\begin{table}[t]
\centering
\small
\begin{tabular}{lcc}
\toprule
\textbf{Training Configuration} & \textbf{Inference Configuration (MoE Routing Top-$k$)} & \textbf{Validation Loss} \\
\midrule
Baseline{\tiny Top-$k=8$} & Top-$k=8$ & 1.945 \\
\midrule
\multirow{4}*{Elastic Sparsity{\tiny Top-$k \in [1, 8]$}} 
& Top-$k=8$ & 1.969 \\
& Top-$k=4$  & 1.971 \\
& Top-$k=2$  & 2.003 \\
& Top-$k=1$  & 2.175 \\
\bottomrule
\end{tabular}
\caption{Validation loss under different routing sparsity levels for the
small-scale MoE model. Elastic sparsity allows stable inference across varying routing budgets, with graceful performance degradation under aggressive sparsification.}
\label{tab:elastic-topk}
\end{table}

\begin{table}[t]
\centering
\footnotesize
\resizebox{\textwidth}{!}{%
\setlength{\tabcolsep}{1.9pt}
\begin{tabular}{lccccccccc}
\toprule
\textbf{Model} & \textbf{AVG.} & \textbf{ZebraLogic} & \textbf{LiveCodeBench v6} & \textbf{TAU2} & \textbf{MMMU} & \textbf{MathVista} & \textbf{VisualPuzzle} & \textbf{SimpleVQA} \\
\midrule
\ourmodel-Exp
& 75.55 & 95.00 & 73.35 & 79.35 & 74.11 & 83.70 & 59.93 & 63.40 \\

\ourmodel-Exp-ES$_{25.0\%}$
& 74.43 & 94.10 & 70.70 & 77.34 & 73.78 & 84.90 & 57.98 & 62.19 \\

\ourmodel-Exp-EA$_{35.8\%}$
& 75.17 & 95.20 & 70.93 & 77.23 & 75.11 & 84.50 & 60.39 & 62.86 \\
\bottomrule
\end{tabular}
}
\caption{Performance comparison between the full \ourmodel-Exp and its elastic variants under reduced routing sparsity (\ourmodel-Exp-ES) and all elastic configurations (\ourmodel-Exp-EA).
Elastic sparsity preserves comparable accuracy with improved decoding efficiency, while the fully elastic model achieves competitive performance despite substantially reduced activated computation and parameter usage.}
\label{tab:ernie-elastic}
\end{table}

\subsubsection{Elastic Training}
\label{sec:elastic-discussion}

Elastic training plays a central role in enabling \ourmodel to adapt to different compute, memory, and latency constraints.
In this section, we analyze elastic training strategies through controlled ablation studies on a small-scale MoE model and report key findings on \ourmodel.
Unless otherwise specified, all elastic configurations are derived from the same pretrained checkpoint.

\paragraph{Controlled-Scale Experiments}
We conduct ablation studies using a small-scale MoE model with 64 experts, 454M activated parameters, and 3.2B total parameters.
The model is trained on 250B tokens with a default routing configuration of top-$k=8$.
Validation loss on held-out data is reported throughout.
Based on this experimental setup, we analyze elasticity along the three dimensions separately.

\begin{itemize}
    \item \textbf{Elastic Depth.}
    We first study elastic depth by varying the number of active transformer
    layers during training.
    Consistent with the configuration of \ourmodel, 80\% of training samples retain the full-depth model, while the remaining 20\% reduced-depth sub-networks.
    As reported in Table~\ref{tab:elastic-depth}, elastic depth slightly improves
    full-depth performance, which indicates a regularization effect introduced by occasional layer dropping.
    Moreover, reduced-depth sub-networks exhibit smooth and predictable performance degradation, suggesting that intermediate representations remain robust under layer removal.
    Overall, elastic depth offers a low-risk and cost-efficient mechanism for deriving multiple deployable sub-models from a single pretrained checkpoint.
    
    \item \textbf{Elastic Width.}
    We next evaluate elasticity over MoE width.
    During training, 80\% of samples retain all 64 experts, while the remaining
    20\% samples configurations with 32 experts.
    Table~\ref{tab:elastic-width} shows that elastic width introduces only a slight degradation at full capacity.
    Although reduced-width configurations incur minor performance fluctuations of larger models, they remain functional without retraining, enabling deployment under strict memory constraints.

    \item \textbf{Elastic Sparsity.}
    We also analyze elastic sparsity by varying routing top-$k$ at inference.
    During training, most instances are equipped with default configuration ($k=8$), while the remaining instances are trained with reduced activated experts.
    As shown in Table~\ref{tab:elastic-topk}, elastic top-$k$ training incurs a modest degradation under the full-activation configuration, while enabling stable and effective inference under substantially reduced routing budgets.
\end{itemize}

\paragraph{Scaling to \ourmodel}
Beyond controlled-scale experiments, we further evaluate the effectiveness of elastic training on the pre-trained experimental model \ourmodel-Exp-Base and its post-trained model \ourmodel-Exp.
Starting from \ourmodel-Exp, we first analyze elastic sparsity by reducing the routing top-$k$ to 25\% during inference.
As summarized in Table~\ref{tab:ernie-elastic}, \ourmodel-Exp-ES$_{25.0\%}$ retains comparable performance across a wide range of text and visual benchmarks. 
The resulting accuracy drop remains minor, while the reduction in routing sparsity brings substantial efficiency gains, providing a \emph{more than 15\% improvement in decoding speed}.
Moreover, we jointly activate elasticity along depth, width, and sparsity, deriving a compact pre-trained model from \ourmodel-Exp-Base, which operates with only \emph{53.7\% of the activated parameters} and \emph{35.8\% of the total parameters}. 
Using the same data and training strategy for mid-training and post-training, we obtain the post-trained model, \ourmodel-Exp-EA$_{35.8\%}$.
Despite its substantially reduced computational footprint, the elastic variant achieves competitive performance across benchmarks, attaining an average score of $75.17$ compared to $75.55$ for the full \ourmodel-Exp.
Moreover, strong robustness on challenging reasoning and perception tasks, such as ZebraLogic and VisualPuzzle, indicates that elastic training effectively mitigates the performance degradation typically associated with aggressive reductions in computation and parameters.

Overall, these results highlight that elasticity is not merely as a post-hoc compression technique, but a principled training paradigm. By jointly optimizing depth, width, and sparsity during pre-training and, the elastic model learns to redistribute representational capacity across layers and modalities, leading to favorable performance–efficiency trade-offs.

\section{Conclusion}

We introduce \ourmodel, a natively unified foundation model that integrates multimodal understanding and generation of text, image, video, and audio with a shared next-group-of-tokens prediction objective.
To the best of our knowledge, \ourmodel represents the first realization of multimodal understanding and generation within a unified, trillion-level autoregressive framework.
An ultra-sparse mixture-of-experts architecture with modality-agnostic expert routing enables scalable cross-modal modeling without relying on modality-specific designs.
We explore a novel elastic training paradigm that supports flexible deployment configurations within a single pre-trained model, which is proved successful to maintain both training efficiency and performances.   
We also address key challenges in reinforcement learning for large-scale multimodal models and ensure pos-training stability and efficiency.
Extensive experiments demonstrate competitive and balanced performance across modalities.
Overall, the results indicate that autoregressive unified multimodal and elastic pre-training provides a scalable pathway toward the next generation of foundational models.

\section{Contributors}

Haifeng Wang, Hua Wu, Tian Wu, Yu Sun, Jing Liu, Dianhai Yu, Yanjun Ma, Jingzhou He, Zhongjun He, Dou Hong, Qiwen Liu, Shuohuan Wang, Junyuan Shang, Zhenyu Zhang, Yuchen Ding, Jinle Zeng, Jiabin Yang, Liang Shen, Ruibiao Chen, Weichong Yin, Siyu Ding, Dai Dai, Shikun Feng, Siqi Bao, Bolei He, Yan Chen, Zhenyu Jiao, Ruiqing Zhang, Zeyu Chen, Qingqing Dang, Kaipeng Deng, Jiajun Jiang, Enlei Gong, Guoxia Wang, Yanlin Sha, Yi Liu, Yehan Zheng, Weijian Xu, Jiaxiang Liu, Zengfeng Zeng, Yingqi Qu, Zhongli Li, Zhengkun Zhang, Xiyang Wang, Zixiang Xu, Xinchao Xu, Zhengjie Huang, Dong Wang, Bingjin Chen, Yue Chang, Xing Yuan, Shiwei Huang, Qiao Zhao, Xinzhe Ding, Shuangshuang Qiao, Baoshan Yang, Bihong Tang, Bin Li, Bingquan Wang, Binhan Tang, Binxiong Zheng, Bo Cui, Bo Ke, Bo Zhang, Bowen Zhang, Boyan Zhang, Boyang Liu, Caiji Zhang, Can Li, Chang Xu, Chao Pang, Chao Zhang, Chaoyi Yuan, Chen Chen, Cheng Cui, Chenlin Yin, Chun Gan, Chunguang Chai, Chuyu Fang, Cuiyun Han, Dan Zhang, Danlei Feng, Danxiang Zhu, Dong Sun, Dongbo Li, Dongdong Li, Dongdong Liu, Dongxue Liu, Fan Ding, Fan Hu, Fan Li, Fan Mo, Feisheng Wu, Fengwei Liu, Gangqiang Hu, Gaofeng Lu, Gaopeng Yong, Gexiao Tian, Guan Wang, Guangchen Ni, Guangshuo Wu, Guanzhong Wang, Guihua Liu, Guishun Li, Haibin Li, Haijian Liang, Haipeng Ming, Haisu Wang, Haiyang Lu, Haiye Lin, Han Zhou, Hangting Lou, Hanwen Du, Hanzhi Zhang, Hao Chen, Hao Du, Hao Liu, Hao Zhou, Haochen Jiang, Haodong Tian, Haoshuang Wang, Haozhe Geng, Heju Yin, Hong Chen, Hongchen Xue, Hongen Liu, Honggeng Zhang, Hongji Xu, Hongwei Chen, Hongyang Zhang, Hongyuan Zhang, Hua Lu, Huan Chen, Huan Wang, Huang He, Hui Liu, Hui Zhong, Huibin Ruan, Jiafeng Lu, Jiage Liang, Jiahao Hu, Jiahao Hu, Jiajie Yang, Jialin Li, Jian Chen, Jian Wu, Jianfeng Yang, Jianguang Jiang, Jianhua Wang, Jianye Chen, Jiaodi Liu, Jiarui Zhou, Jiawei Lv, Jiaxin Zhou, Jiaxuan Liu, Jie Han, Jie Sun, Jiefan Fang, Jihan Liu, Jihua Liu, Jing Hu, Jing Qian, Jing Yan, Jingdong Du, Jingdong Wang, Jingjing Wu, Jingyong Li, Jinheng Wang, Jinjin Li, Jinliang Lu, Jinlin Yu, Jinnan Liu, Jixiang Feng, Jiyi Huang, Jiyuan Zhang, Jun Liang, Jun Xia, Jun Yu, Junda Chen, Junhao Feng, Junhong Xiang, Junliang Li, Kai Liu, Kailun Chen, Kairan Su, Kang Hu, Kangkang Zhou, Ke Chen, Ke Wei, Kui Huang, Kun Wu, Kunbin Chen, Lei Han, Lei Sun, Lei Wen, Linghui Meng, Linhao Yu, Liping Ouyang, Liwen Zhang, Longbin Ji, Longzhi Wang, Meng Sun, Meng Tian, Mengfei Li, Mengqi Zeng, Mengyu Zhang, Ming Hong, Mingcheng Zhou, Mingming Huang, Mingxin Chen, Mingzhu Cai, Naibin Gu, Nemin Qiu, Nian Wang, Peng Qiu, Peng Zhao, Pengyu Zou, Qi Wang, Qi Xin, Qian Wang, Qiang Zhu, Qianhui Luo, Qianwei Yang, Qianyue He, Qifei Wu, Qinrui Li, Qiwen Bao, Quan Zhang, Quanxiang Liu, Qunyi Xie, Rongrui Zhan, Rufeng Dai, Rui Peng, Ruian Liu, Ruihao Xu, Ruijie Wang, Ruixi Zhang, Ruixuan Liu, Runsheng Shi, Ruting Wang, Senbo Kang, Shan Lu, Shaofei Yu, Shaotian Gong, Shenwei Hu, Shifeng Zheng, Shihao Guo, Shilong Fan, Shiqin Liu, Shiwei Gu, Shixi Zhang, Shuai Yao, Shuang Zhang, Shuangqiao Liu, Shuhao Liang, Shuwei He, Shuwen Yang, Sijun He, Siming Dai, Siming Wu, Siyi Long, Songhe Deng, Suhui Dong, Suyin Liang, Teng Hu, Tianchan Xu, Tianliang Lv, Tianmeng Yang, Tianyi Wei, Tiezhu Gao, Ting Sun, Ting Zhang, Tingdan Luo, Wei He, Wei Luan, Wei Yin, Wei Zhang, Wei Zhou, Weibao Gong, Weibin Li, Weicheng Huang, Weichong Dang, Weiguo Zhu, Weilong Zhang, Weiqi Tan, Wen Huang, Wenbin Chang, Wenjing Du, Wenlong Miao, Wenpei Luo, Wenquan Wu, Xi Shi, Xi Zhao, Xiang Gao, Xiangguo Zhang, Xiangrui Yu, Xiangsen Wang, Xiangzhe Wang, Xianlong Luo, Xianying Ma, Xiao Tan, Xiaocong Lin, Xiaofei Wang, Xiaofeng Peng, Xiaofeng Wu, Xiaojian Xu, Xiaolan Yuan, Xiaopeng Cui, Xiaotian Han, Xiaoxiong Liu, Xiaoxu Fei, Xiaoxuan Wu, Xiaoyu Wang, Xiaoyu Zhang, Xin Sun, Xin Wang, Xinhui Huang, Xinming Zhu, Xintong Yu, Xinyi Xu, Xinyu Wang, Xiuxian Li, XuanShi Zhu, Xue Xu, Xueying Lv, Xuhong Li, Xulong Wei, Xuyi Chen, Yabing Shi, Yafeng Wang, Yamei Li, Yan Liu, Yanfu Cheng, Yang Gao, Yang Liang, Yang Wang, Yang Wang, Yang Yang, Yanlong Liu, Yannian Fu, Yanpeng Wang, Yanzheng Lin, Yao Chen, Yaozong Shen, Yaqian Han, Yehua Yang, Yekun Chai, Yesong Wang, Yi Song, Yichen Zhang, Yifei Wang, Yifeng Guo, Yifeng Kou, Yilong Chen, Yilong Guo, Yiming Wang, Ying Chen, Ying Wang, Yingsheng Wu, Yingzhan Lin, Yinqi Yang, Yiran Xing, Yishu Lei, Yixiang Tu, Yiyan Chen, Yong Zhang, Yonghua Li, Yongqiang Ma, Yongxing Dai, Yongyue Zhang, Yu Ran, Yu Sun, Yu-Wen Michael Zhang, Yuang Liu, Yuanle Liu, Yuanyuan Zhou, Yubo Zhang, Yuchen Han, Yucheng Wang, Yude Gao, Yuedong Luo, Yuehu Dong, Yufeng Hu, Yuhui Cao, Yuhui Yun, Yukun Chen, Yukun Gao, Yukun Li, Yumeng Zhang, Yun Fan, Yun Ma, Yunfei Zhang, Yunshen Xie, Yuping Xu, Yuqin Zhang, Yuqing Liu, Yurui Li, Yuwen Wang, Yuxiang Lu, Zefeng Cai, Zelin Zhao, Zelun Zhang, Zenan Lin, Zezhao Dong, Zhaowu Pan, Zhaoyu Liu, Zhe Dong, Zhe Zhang, Zhen Zhang, Zhengfan Wu, Zhengrui Wei, Zhengsheng Ning, Zhenxing Li, Zhenyu Li, Zhenyu Qian, Zhenyun Li, Zhi Li, Zhichao Chen, Zhicheng Dong, Zhida Feng, Zhifan Feng, Zhihao Deng, Zhijin Yu, Zhiyang Chen, Zhonghui Zheng, Zhuangzhuang Guo, Zhujun Zhang, Zhuo Sun, Zichang Liu, Zihan Lin, Zihao Huang, Zihe Zhu, Ziheng Zhao, Ziping Chen, Zixuan Zhu, Ziyang Xu, Ziyi Liang, Ziyuan Gao

% \clearpage
% \appendix
% \input{latex_files/content/appendix.tex}

\clearpage
\bibliography{biblio}
\bibliographystyle{colm2024_conference}

\end{document}